%% file: iclr2026_conference.tex
\documentclass{article} % For LaTeX2e
\usepackage{iclr2026_conference,times}

\input{math_commands.tex}

\usepackage{hyperref}
\usepackage{url}

\usepackage{booktabs}
\usepackage{multirow}
\usepackage{array}
\usepackage{makecell}
\usepackage{graphicx} 
\usepackage{hyperref}
\usepackage{xcolor,colortbl} 
\hypersetup{
    colorlinks=true,  % Activates colored links instead of boxes
    citecolor=blue,   % Sets the color for citations
    linkcolor=blue,    % Sets the color for internal links (e.g., table of contents)
    urlcolor=cyan     % Sets the color for URLs
}
\usepackage[utf8]{inputenc}
\usepackage{amsmath}
\usepackage[ruled,vlined]{algorithm2e}
\usepackage{wrapfig}
\usepackage{algorithm2e}
\usepackage{graphicx}
\usepackage{float} % 提供 [H]
\usepackage{subcaption}
\usepackage{adjustbox}
\usepackage{listings}
\usepackage{xcolor}
\definecolor{diffgreen}{RGB}{0, 128, 0}
\definecolor{diffred}{RGB}{220, 50, 47}
\definecolor{diffbggreen}{RGB}{230, 255, 230}
\definecolor{diffbgred}{RGB}{255, 230, 230}
\definecolor{commentcolor}{RGB}{106, 153, 85}
\definecolor{keywordcolor}{RGB}{0, 0, 255}
\definecolor{stringcolor}{RGB}{163, 21, 21}
\newcommand{\AlgoSize}{\normalsize}
\SetAlFnt{\AlgoSize}
\SetAlCapFnt{\AlgoSize}
\SetAlCapNameFnt{\AlgoSize}

\newcommand{\zhangcong}[1]{\textcolor{black}{#1}}

\title{Refining Hybrid Genetic Search for CVRP via Reinforcement Learning-Finetuned LLM}

\author{
Rongjie Zhu$^{1,}$\thanks{Authors contributed equally}   , 
Cong Zhang$^{2,}$\footnotemark[1]   ,
Zhiguang Cao$^{3}$\\
$^{1}$School of Teacher Education, Nanjing University of Information Science and Technology, China\\
$^{2}$Nanyang Technological University, Singapore\\
$^{3}$Singapore Management University, Singapore\\
\texttt{rzhu114514@gmail.com, cong.zhang92@gmail.com}\\
\texttt{zhiguangcao@outlook.com}
}

\iclrfinalcopy % Uncomment for camera-ready version, but NOT for submission.
\begin{document}

% 写作：
% 1. highlight 训练没超过400个点
% 2. 第一个小模型改进大框架，其余的无一不是用强大模型通过复杂的链路设计和prompt engineer达成的，我们是第一个端到端
% 3. 大多数NCO的方法并没有测这么大size的，我们尝试了他们的方法，不是因为OOM跑不了，要不就是很慢，要不就没有提供跑cvrplib的方法，并且很少有论文测这么大size
% 4. 更改 abstract，强调 size 和 iter 的泛化效果
% 5. 现在基于大模型的工作 optimal 都太差了，我们是第一个用小模型去改进成熟框架的，效果非常好
% 6. 表格中的结果，不论是GPT的模型还是我们的模型，我们都是采样16次，取最好的，如果16次采样的结果都不可用，那就是 NA
% 7. GPT 都是改一些无关紧要的是吧，我们的改动的少，但是关键，有保留可读性

% 写作：
% 1. highlight 训练没超过400个点
% 2. 第一个小模型改进大框架，其余的无一不是用强大模型通过复杂的链路设计和prompt engineer达成的，我们是第一个端到端
% 3. 大多数NCO的方法并没有测这么大size的，我们尝试了他们的方法，不是因为OOM跑不了，要不就是很慢，要不就没有提供跑cvrplib的方法，并且很少有论文测这么大size
% 4. 更改 abstract，强调 size 和 iter 的泛化效果
% 5. 现在基于大模型的工作 optimal 都太差了，我们是第一个用小模型去改进成熟框架的，效果非常好
% 6. 表格中的结果，不论是GPT的模型还是我们的模型，我们都是采样16次，取最好的，如果16次采样的结果都不可用，那就是 NA
% 7. 我们尝试用 reasoning 模型 + 强化学习来做
% 8. 学习模式的分析，可以看模型是否是先学会写合法的算子->写成功的算子->写效果更好的算子，按照这个顺序有阶梯的去学习
% 9. ab highlight 比4o好

\maketitle

\begin{abstract}
While large language models (LLMs) are emerging as automated heuristic designers for solving vehicle routing problems (VRPs), state-of-the-art approaches predominantly rely on massive, general-purpose models like GPT-4. This work challenges this paradigm by demonstrating that smaller, specialized LLMs, when finely tuned, can generate components that surpass expert-designed heuristics within advanced solvers. We introduce RFTHGS, a novel \textbf{\underline{R}}einforcement learning (RL) framework for \textbf{\underline{F}}ine-\textbf{\underline{T}}uning a small LLM to produce high-performance crossover operators for the \textbf{\underline{H}}ybrid \textbf{\underline{G}}enetic \textbf{\underline{S}}earch (HGS) solver to solve the capacitated vehicle routing problem (CVRP). Our methods utilizes a multi-tiered, curriculum-based reward function that progressively guides the LLM to first produce compilable code, then executable operators, and finally, components that exceed human expert-designed ones. Additionally, we introduce an operator caching mechanism to work in conjunction with the reward function, discouraging plagiarism and promoting diversity during training. Experimental results demonstrate that our fine-tuned LLM generates crossover operators which significantly outperform those designed by human experts in HGS. This performance advantage is consistent, holding from small-scale instances and generalizing to large-scale problems of up to 1000 nodes. Furthermore, RFTHGS surpasses leading neuro-combinatorial baselines, prompt-based methods, and commercial LLMs, including GPT-4o and GPT-4o-mini.

\end{abstract}
% for the well-known CVRPLIB benchmarks

\section{Introduction}
Combinatorial Optimization Problems (COPs) represent a fundamental class of computational challenges that arise across diverse domains including supply chain management, logistics, scheduling, and network design~\citep{bengio2021machine}. These problems, characterized by their discrete decision variables and complex constraints, are often NP-hard in complexity, making exact solutions computationally intractable for large-scale instances. For decades, researchers have developed specialized algorithms, heuristics, and metaheuristics to approximate optimal solutions, yet these approaches typically require significant domain expertise and manual design efforts~\citep{papadimitriou1998combinatorial}. The emergence of large language models (LLMs), with their remarkable reasoning and pattern recognition capabilities, has introduced a transformative paradigm for tackling COPs. By leveraging their natural language processing and generative abilities, LLMs enable automated approaches that reduce reliance on manual algorithm design and expert intervention~\citep{sun2024autosat,novikov2025alphaevolve}. Investigating on leveraging LLMs to solve Vehicle Routing Problems (VRPs) represents one of the most cutting-edge frontiers in this field.

Initial studies explored the use of LLMs as end-to-end solvers for VRPs~\citep{yang2024large}. However, these purely generative approaches often yield solutions that are substantially inferior to those from conventional or deep learning-based solvers and are frequently infeasible, i.e., a shortcoming attributed to the fact that LLMs are notoriously prone to hallucinations~\citep{kalai2025languagemodelshallucinate}. Consequently, a more promising direction is to integrate LLMs not as autonomous solvers but as intelligent operators within established optimization frameworks, such as evolutionary algorithms. In this hybrid paradigm, the LLM acts as a strategic generator or refiner within an iterative loop. For instance, \citet{10611913} employ general-purpose LLMs  to guide an evolutionary process, using in-context prompting to perform crossover and mutation. An alternative approach inverts this relationship, using evolutionary computation not as a framework for the LLM to power, but as a mechanism to guide the LLM itself in generating increasingly effective heuristics. Representative works like EoH~\citep{10.5555/3692070.3693374} and ReEvo~\citep{NEURIPS2024_4ced59d4} iteratively refine LLM-generated heuristic through evolutionary selection. 
% In contrast, a distinct research direction prioritizes general-purpose frameworks designed for diverse VRP variants, emphasizing generalization applicability while sacrificing certain performance. 
In contrast, another line of research prioritizes general-purpose frameworks for diverse VRP variants, favoring broader generalization at the cost of performance.
Methods such as ARS~\citep{li2025ars}, which leverage predefined structures to generate constraint-checking functions, and DRoC~\citep{jiang2025droc}, which utilizes retrieval to produce code that invokes external solvers like OR-Tools~\citep{ortools_routing}, demonstrate improved generalization and robustness against code execution failures. Nonetheless, a significant performance gap remains with conventional and deep learning-based solvers, echoing doubts about the immediate application of existing methods to large-scale problems. Given that practical instances in the real scenario often rely on advanced solvers, a critical open question is whether we can finetune small LLMs to optimize key components within these solvers to achieve beyond expert performance, presenting a challenging yet promising research frontier.

We introduce RFTHGS, a reinforcement learning (RL) framework that fine-tunes a reasoning LLM with 14B parameters to autonomously generate effective crossover operators for the Hybrid Genetic Search (HGS) algorithm~\citep{vidal2022hybrid}, thereby enhancing its performance on large-scale Capacitated Vehicle Routing Problems (CVRP). The framework leverages solution quality as the principal feedback signal, realized through a structured, tiered reward scheme designed to guide the learning process across three progressive stages. First, since instruction-tuned models often struggle to produce syntactically valid code, we assign an initial reward for generating compilable code. Second, an additional reward is granted when the generated operator executes successfully without runtime errors or timeouts. Finally, for compilable and executable operators, a linear reward is assigned based on their relative improvement in solution quality over a baseline expert-designed operator, measured on a predefined set of CVRP instances. These instances are used solely for evaluating generated operators and are never exposed to the model during generation, ensuring fairness and generalization. To mitigate reward hacking and discourage repetitive outputs, we incorporate an operator buffer mechanism that penalizes duplicate generations, thereby explicitly promoting diversity among discovered operators. Through this iterative refinement process, RFTHGS enables the LLM to evolve crossover operators that ultimately surpass handcrafted ones designed by human experts. Extensive experiments demonstrate that the LLM-generated operator delivers substantial improvements over the expert-designed counterpart in HGS, achieving superior performance on both small and large-scale benchmarks (up to 1,000 nodes). Moreover, it consistently outperforms leading neuro-combinatorial and prompt-based LLM baselines by a significant margin. To the best of our knowledge, this work provides the first empirical evidence that a compact reasoning LLM (14B parameters) can be fine-tuned via RL to generate critical algorithmic components that exceed the performance of those in state-of-the-art, expert-engineered solvers.

\section{Related Work}
\subsection{On The Reasoning Ability of Large Language Models}
The development of large language models (LLMs) with advanced reasoning capabilities has evolved through several key phases, beginning with Chain of Thought (CoT) prompting, which explicitly guides models to generate intermediate reasoning steps, significantly improving performance on tasks like arithmetic and commonsense reasoning~\citep{plaat2024reasoning,wei2022chain}. This approach was further enhanced by inference-time strategies such as Self-Consistency (aggregating multiple reasoning paths) and Tree-of-Thoughts (ToT) (exploring branched reasoning trajectories), which reduce errors and improve robustness in multi-step problem-solving~\citep{yao2023tree}. A major shift occurred with the integration of reinforcement learning (RL) techniques, where models are trained using verifiable rewards (e.g., correct answers in mathematical problems or code execution results) to incentivize logical reasoning without relying solely on supervised fine-tuning~\citep{xiang2025towards,xu2025towards}. For instance, DeepSeek-R1~\citep{deepseekai2025deepseekr1incentivizingreasoningcapability} and OpenAI's o1 series~\citep{openai2024openaio1card} exemplify how RL-driven self-improvement and scaled inference-time compute enable deliberate, step-by-step reasoning. Additionally, hybrid methods such as Microsoft's rStar-Math~\citep{guan2025rstar} (which integrates Monte Carlo Tree Search for problem decomposition) and retrieval-augmented generation (RAG)~\citep{NEURIPS2020_6b493230} combine the pattern recognition capabilities of LLMs with external tools for rigorous symbolic operations. Recent advancements also focus on test-time training and outcome-based exploration to enhance adaptability and diversity in reasoning paths, while interpretability research aims to ensure faithful internal reasoning processes~\citep{song2025outcomebasedexplorationllmreasoning}. Despite progress, challenges such as hallucination, scalability, and generalisation persist, driving ongoing innovation in architectures and training paradigms~\citep{shojaee2025illusionthinkingunderstandingstrengths}. In contrast to the well-recognized success in solving math problems, training the LLM with reasoning capabilities to generate operators that can outperform the default expert-designed ones in advanced VRP solvers remains challenging and largely unexplored.

\subsection{Solving CVRP With LLM In The Loop}
The integration of LLMs into Vehicle Routing Problems (VRPs) has primarily advanced through prompting-based methodologies~\citep{yang2024large,jiang2025droc,10611913,huang2024large}, which leverage the robust reasoning capabilities of state-of-the-art (SOTA) off-the-shelf LLMs (e.g., GPT-o3-mini). For instance, ARS~\citep{li2025arsautomaticroutingsolver} uses LLMs to automatically generate constraint-aware heuristics for solving complex vehicle routing problems by synthesizing natural language descriptions into executable code, which can construct heuristics for 90\% of common VRP variants with different constraints. Similarly, Hercules~\citep{wu2025efficient} employs Core Abstraction Prompting (CAP) to derive high-performance heuristics by abstracting core components from elite solutions, though it remains dependent on powerful closed-source LLMs. Existing research is also investigating automatic heuristic design with LLM. Representative works like EoH~\citep{10.5555/3692070.3693374} and ReEvo~\citep{NEURIPS2024_4ced59d4} iteratively refine LLM-generated heuristics through evolutionary selection. Most recently, CALM~\citep{huang2025calmcoevolutionalgorithmslanguage} extends this paradigm by integrating reinforcement fine-tuning of the LLM into the evolutionary loop, allowing the model and its generated heuristics to co-evolve. In contrast, finetuning-based approaches for COP remain relatively sparse, often due to challenges like catastrophic forgetting, computational costs, and overfitting when adapting pre-trained models to specialized domains. While parameter-efficient methods like LoRA~\citep{Hu2021LoRALA} mitigate some issues, fine-tuning small open-source LLMs (e.g., LLaMA~\citep{Touvron2023LLaMAOA}) to generate operators for widely-used solvers like Hybrid Genetic Search (HGS)~\citep{vidal2022hybrid} remains an open problem. Current efforts are mainly devoted to prompting, leaving a gap in developing specialized, lightweight models that can efficiently integrate with solver frameworks without relying on API-dependent, proprietary LLMs.

\section{Preliminary}

\textbf{HGS For Solving The Capacitated Vehicle Routing Problem (CVRP)}.
The Hybrid Genetic Search (HGS)~\citep{vidal2022hybrid} algorithm represents a state-of-the-art metaheuristic framework prominently applied to complex combinatorial optimization problems, particularly vehicle routing problems (VRPs). As an extension of the classical genetic algorithm, HGS distinguishes itself through a tight integration of population-based evolutionary search and intensive local improvement procedures. Its core mechanism involves maintaining a diverse population of solutions that are iteratively refined through a process of selection, crossover, and local search. Within this framework, the crossover operator is the primary mechanism for global exploration by recombining genetic material from parent solutions to generate novel offspring. This operator does not merely produce trivial combinations; rather, it constructs promising, high-quality solution skeletons that effectively inherit desirable attributes from both parents. These offspring solutions subsequently undergo rigorous local search, which acts upon the foundation laid by crossover to exploit the solution space locally and achieve feasibility and optimality. This synergistic interplay, where crossover provides a robust starting point for deep local exploitation, is a critical factor in the documented efficacy of HGS, enabling it to navigate the trade-off between exploration and exploitation effectively and consistently produce high-quality solutions for routing problems.

\textbf{The Group Relative Policy Optimization (GRPO) Algorithm}.
% \textbf{}. Group Relative Policy Optimization 
GRPO~\citep{shao2024deepseekmathpushinglimitsmathematical} is an improved variant of Proximal Policy Optimization (PPO)~\citep{schulman2017proximal}. The key innovation of GRPO lies in utilizing a normalized reward function to compute advantages, where the mean and variance are estimated through Monte-Carlo sampling (with sample size $G$) from the current policy $\pi_k(\cdot|x)$ at step $k$ for each input (prompt) $x$. For given parameters $\epsilon$, $\beta > 0$, and a reference policy $\pi_{ref}$ (usually the base model), the GRPO objective optimization problem is formulated as:

\begin{equation}
\max_\pi \mathbb{E}_{y\sim\pi_k(\cdot|x)} \min \left[ \frac{\pi(y|x)}{\pi_k(y|x)} A_{\pi_k}(x, y), \text{clip} \left( \frac{\pi(y|x)}{\pi_k(y|x)}, 1-\varepsilon, 1+\varepsilon \right) A_{\pi_k}(x, y) \right] - \beta\text{KL}(\pi||\pi_{\text{ref}})
\label{eq:grpo_objective}
\end{equation}

where KL denotes Kullback-Leibler divergence, and $A_{\pi_k}$ represents GRPO advantage function:

\begin{equation}
A_{\pi_k}(x, y_i) = \frac{r(x, y_i) - \mathbb{E}_{\pi_k} r(x, y_i)}{\sqrt{\mathbb{E}_{\pi_k}(r(x, y_i) - \mathbb{E}_{\pi_k} r(x, y_i))^2 + \varepsilon}} \simeq \frac{r(x, y_i) - \mu(\{r_\ell\})}{\sqrt{\sigma^2(\{r_\ell\}) + \varepsilon}}, \quad 1 \leq \ell \leq G
\label{eq:advantage_estimation}
\end{equation}

with the advantage estimated by sampling a "group" of size $G$ for each input $x$, and $\mu$ and $\sigma$ represent the empirical mean and standard deviation, respectively. 

\begin{figure}[t]
    \centering
    % \vspace{5pt}
    \includegraphics[width=1.05\textwidth]{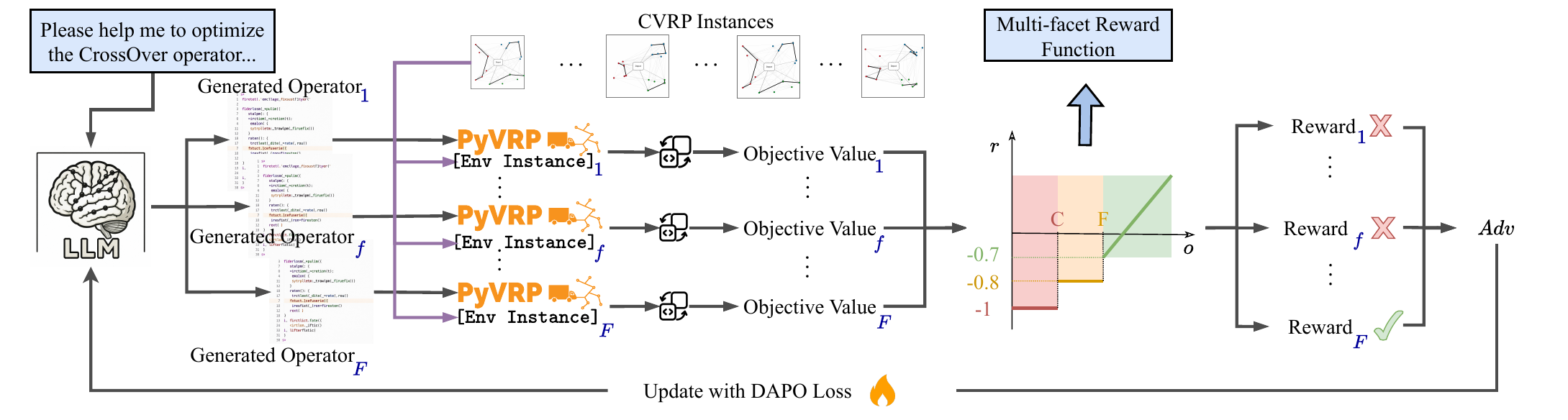}
    \caption{\textbf{The reinforcement learning pipeline of RFTHGS.} The framework iteratively optimizes an LLM to generate effective crossover operators for HGS. Each iteration consists of generating code from a structured prompt, evaluating the operator's performance on a validation set (using incremental compilation for speed), calculating a multi-faceted reward, and updating the LLM policy. The LLM only sees operator examples, not problem instances or the solver codebase.}
    % \vspace{-1em}
    \label{fig:RFTHGS}
\end{figure}

\section{Method}
We introduce RFTHGS, a reinforcement learning framework that fine-tunes large language models (LLMs) to generate crossover operators that outperform expert-designed ones in the Hybrid Genetic Search (HGS) solver~\citep{vidal2022hybrid}. The framework uses solution quality as the key reward signal to guide the LLM toward generating increasingly effective operators. As shown in Figure~\ref{fig:RFTHGS}, RFTHGS is an iterative closed feedback loop. Each iteration consists of prompting the LLM to generate new operators, assessing their performance on a set of predefined CVRP instances, and then employing the feedback reward to refine the LLM via reinforcement learning.

Specifically, each iteration begins by constructing a few-shot CoT context (see Appendix~\ref{prompt}) that contains: (1) instructions specifying key properties (e.g., diversity and quality) and steps for generating a high-quality operator, and (2) examples of existing operators that illustrate the required structure and syntax. The LLM then generates new crossover operators based exclusively on this prompt, with no references to other modules in the HGS library or access to specific CVRP instances. For evaluation, each generated operator is integrated into the HGS library, and the code is recompiled to test on a fixed problem benchmark set. We employ incremental compilation to speed up this recompilation step. Finally, performance metrics such as compilability and improvement over baseline operators are combined into a multi-faceted reward, which is used to finetune the LLM via reinforcement learning. The RFTHGS framework automates the design of optimization operators, and demonstrates that small LLMs can evolve components that outperform human-designed ones in a state-of-the-art CVRP solver.

\subsection{One-Step POMDP Modelling}
We formulate the operator optimization as a one-step Partially Observable Markov Decision Process (One-Step POMDP), with formal definitions of state, action, reward and policy given as follows. 
% We give the formal definition of the POMDP as follows.

% \begin{wrapfigure}{r}{0.5\textwidth}
%     \centering
%     \includegraphics[width=0.5\textwidth]{fig_reward_fn.pdf}
%     \vspace{-2em}
%     \caption{The \zhangcong{multi-faceted} reward function.}
%     \label{fig:fig_reward_fn}
% \end{wrapfigure}

\begin{figure}[t]
    \centering
    \begin{subfigure}[b]{0.56\textwidth}
        \centering
        \includegraphics[width=\textwidth]{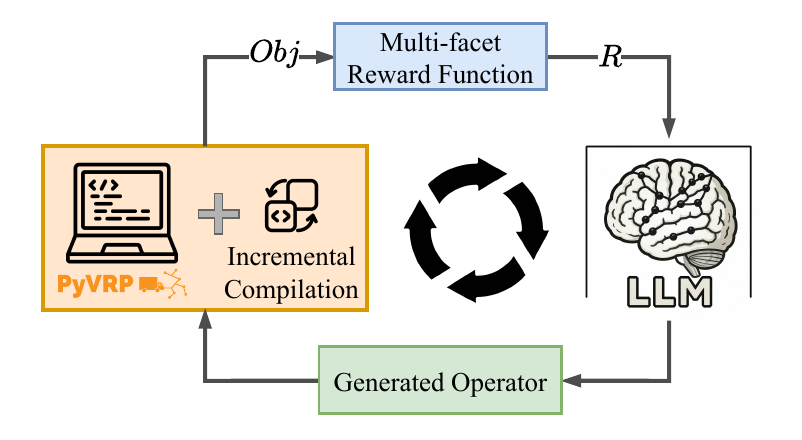}
        \caption{}
        \label{fig:fig_HGS_as_Env}
    \end{subfigure}
    \hfill
    \begin{subfigure}[b]{0.42\textwidth}
        \centering
        \includegraphics[width=\textwidth]{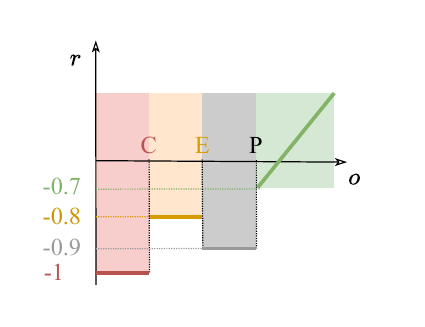}
        \caption{}
        \label{fig:fig_reward_fn}
    \end{subfigure}
    \caption{(a) HGS as the environment for evaluating the quality of LLM-generated operators. We use the incremental compilation technique to boost the computation of objective values. (b) The multi-faceted reward function.}
    % \vspace{-1.5em}
    \label{fig:overview}
\end{figure}

\textbf{State}: % we only have initial state, and the state is changing due to the anti-plagiarism buffer. no state transit. 这是一个POMDP
The state $X \in \mathcal{X}$\footnote{Here we use the symbol $X \in \mathcal{X}$ to represent the state to highlight that the RL task here is different from the conventional RL ones, where the initial state is the input to the LLM.} denotes a tokenized prompt received by the LLM, comprising both the task instructions and examples of the target operator to be optimized. To mitigate context overload, the input is restricted to the target operator itself (e.g., the crossover operator in HGS), rather than the entire library of the solver. This restriction results in a partially observable environment, as the model only perceives a subset of the full state space (i.e., the complete solver repository). 

To enhance the versatility of the learning process, we maintain a buffer of few-shot examples containing operators generated by the LLM during training as well as those designed by human experts. At each iteration, we randomly sample examples from this buffer to construct the prompt. This approach enables the LLM to learn from and attempt to improve upon both its previous generations and expert-designed operators. Furthermore, it enriches the diversity of the initial states (i.e., prompts), which helps prevent overfitting and encourages broader exploration.

\textbf{Action}: % generated operator, 给一个 a 的logp 展示一下
% The action $Y \in \mathcal{Y}$ given by our operator-refining LLM is a series of tokens consisting of two main parts. The first part is the reasoning from LLM, summerizing the steps and corresponding rationales the LLM plans to approach the optimization task, enclosed by the special token \texttt{<think>} and \texttt{<$\backslash$think>}. The second half of the action is the generated operator code. 
% Here since we are optimizing the 
The action (or response) $Y \sim p(\cdot | X \in \mathcal{X})$ generated by our operator-refining LLM is a sequence of tokens $Y  \in \mathcal{Y}$ consisting of two main parts, where $p(\cdot | X)$ is the conditional probability from which the actions are sampled. The first part is a reasoning segment enclosed between the special tokens \texttt{<think>} and \texttt{</think>}, where it explains the steps and logic it plans to take for the optimization task. Following this, it outputs optimized version of the code for the target operator.

\textbf{State Transition}: The state terminates after the action is generated, since we only allow one round of optimization for the operator. Therefore, there are no state transitions, and the POMDP is one-step.

\textbf{Reward}: % HGS as env, incremental complilation
The reward $r \in \mathbb{R}$ is a scalar evaluating the quality of LLM-generated operators. Please refer to Section~\ref{reward_design} below for details of the reward function.

% We designed a multi-step reward function to decompose the learning process into manageable stages, thereby incrementally guiding the LLM to generate operators that surpass expert-designed ones. 

\textbf{Policy Network}: % LLM, 给出 p(.|x) = ...
% The policy network is a base large language model  with trainable parameters $\theta$ that approximate the conditional probability from which we can sample the optimized operator. For the policy network, we are mainly referring to those small (7B or 14B) LLMs with open-source weights. The policy can be a pretrained base model or an instruction-finetuned one. In this work, we adopt the Qwen3-14B as the policy network, and build out operator refining policy on top of it.
The policy network is a base large language model, denoted by $\pi_\theta(\cdot | X),\ X \in \mathcal{X}$, with trainable parameters $\theta$ that parameterize the conditional probability distribution $p(\cdot | X)$ from which the optimized operator is sampled. In this work, we focus on relatively small LLMs (e.g., 14B parameters), which can be either a pretrained base model or an instruction-tuned variant.

% \citep{yang2025qwen3}

% \begin{wrapfigure}{r}{0.5\textwidth}
%     \centering
%     \includegraphics[width=0.5\textwidth]{fig_HGS_as_Env.pdf}
%     \caption{HGS as the environment for evaluating the quality of LLM-generated operators. We use the incremental compilation technique to boost the computation of objective values.}
%     \label{fig:fig_HGS_as_Env}
% \end{wrapfigure}

% like a bandit, single-step RL

\subsection{The Multi-Faceted Reward Design With Anti-Plagiarism Cache}\label{reward_design}

Building on the insight that carefully crafted, multi-faceted rewards are crucial for effective RL~\citep{narvekar2020curriculum,eppe2022intelligent,huang2025calmcoevolutionalgorithmslanguage}, we developed a multi-tiered reward function to decompose the learning process. Particularly, the reward function follows a curriculum learning principle, guiding the LLM through progressive stages to evolve operators that exceed those designed by human experts. To ensure the robustness of this approach, we further introduce two key innovations: a mechanism to prevent reward hacking by deterring plagiarism of prompt examples, and a method to significantly accelerate the training process.

\textbf{Anti-Plagiarism Cache With Abstract Syntax Tree}.
To mitigate reward hacking and encourage the exploration of unseen operators, we introduce a caching mechanism that leverages Abstract Syntax Trees (ASTs) to deter plagiarism. The AST provides a structured, hierarchical representation that abstracts away unnecessary syntactic details like punctuation and formatting to capture the essential logical structure of the generated operators. We cache the AST representations of all few-shot operator examples in the prompt. For each operator generated by the policy $\pi_\theta$, its AST is compared against those in the cache. A penalty is invoked by the reward function if a substantial match is detected, indicating direct copying. This approach promotes diverse exploration by penalizing redundant operator generation. 
% An ablation study in Section~\ref{AST_cache} empirically validates the efficacy of this anti-plagiarism cache.

% for each generation of LLM-generated operators $o_{i}$, we compare

\textbf{HGS With Incremental Compilation As The Evaluator}.\label{HGS_env}
We have to integrate each generated operator into the HGS library to evaluate its quality. This process inevitably requires recompiling the repository, which will incur prohibitive computational overhead, especially for large training batch sizes. Nonetheless, recompiling the entire library is unwarranted when only a single, small code snippet (the generated operator) is modified. To address this bottleneck, we employ an incremental compilation technique that selectively recompiles only the modified code and its dependencies, reducing recompilation time to approximately 25\% of compiling the whole library and significantly accelerating the training speed. 

% The reward function operates in three progressive stages.
Here we give the formulation of our three-stage reward function.
First, the reward function assigns a reward of $-0.8$ for a syntactically correct and compilable operator to encourage a rapid transition from invalid code and improve exploration efficiency, or a penalty of $-1$ for invalid output. Upon achieving compilability, the function then assesses executability, penalizing runtime failures such as timeouts. 
An operator that executes successfully receives a reward of $-0.7$, independent of its solution quality.
% For the executable operators, they will receive a $-0.7$ reward no matter the performance of the operator. 
Finally, for executable operators, performance is evaluated on a predefined set of CVRP instances, with the reward quantified as the relative improvement over expert-designed benchmarks according to the following calculation:
% \vspace{-5em}

\begin{equation}
r(o) = \begin{cases}
-1 & o \notin \text{C} \\
-0.8 & o \in \text{C}, o \notin \text{E}\\
-0.9 & o \in \text{C}, o \in \text{E},o \in \text{P}\\
\max\left(-0.7, [\phi^J_{\text{HGS}}(o_\text{expert}) - \phi^J_{\text{HGS}}(o)]/\phi^J_{\text{HGS}}(o_{\text{expert}})\right) & o \in \text{C}, o \in \text{E}
, o \notin \text{P}
\end{cases}
\label{eq:multi_faceted_reward}
\end{equation}
% \vspace{-2em}

In this formulation, $o$ represents the generated operator, while C, E, and P correspond to the sets of compilable, executable, and plagiarized code, respectively. For evaluation, the HGS library is recompiled to include the generated operator $o$. The performance metric $\phi^J_{\text{HGS}}(\cdot)$ is then calculated as the average result on $J$ random CVRP instances. To benchmark the effectiveness of our continuous reward design in Equation~\ref{eq:multi_faceted_reward}, we compare it with a discrete version where we employ a +1 reward if the generated operator outperforms the baseline operator in Table~\ref{tab:ablation_reward_design_cp} (details in Appendix~\ref{reward_ablation}). We find that our continuous reward offers feedback proportional to performance gains, enabling sustained refinement and explaining its superior performance.

% where $o$, C, E, and P denote the generated operator, the set of compilable code, the set of executable code, and the set of plagiarized code, respectively. $\phi_{HGS}$ represents the evaluated results of generated operator $o$ with HGS library. Here we evaluate on 30 randomly sampled CVRP instances and take the average.

\input{table_cvrplib_result}
\vspace{-5pt}

\subsection{The Reinforcement Learning Algorithm}
We use DAPO~\citep{yu2025dapoopensourcellmreinforcement} as the reinforcement learning algorithm for training our operator refining network. Specifically, DAPO is an improved version of GRPO with four adjustments: 1). \textit{\underline{Clip-Higher Mechanism}}. Unlike GRPO following the original PPO setting where a unified clip ration is adopted for the positive and negative responses, DAPO decouples the clipping range into a higher upper bound ($\varepsilon_{\text{high}}$) and a standard lower bound ($\varepsilon_{\text{low}}$), allowing the policy to more aggressively increase probabilities for promising but initially low-likelihood tokens. This promotes greater exploration and diversity in generated responses, effectively preventing entropy collapse where the model becomes overly deterministic; 2). \textit{\underline{Dynamic Sampling}}. This strategy filters out prompt groups where all sampled responses are either all correct or all incorrect, as these yield zero advantage and provide no learning signal. By replacing them with new prompts that exhibit varied performance, DAPO ensures every training batch contains meaningful gradients, improving training efficiency and stability without sacrificing throughput. However, in our paper, we deprecate this design as the reward signal in our case is continuous, specifying a wide range of situations from uncompilable code to superior performance gain against the baseline operators that all contribute useful learning signals for the LLM to learn; 3). \textit{\underline{Token-Level Policy Gradient Loss}}. Unlike GRPO, which averages losses at the response level, DAPO calculates and aggregates the loss over all tokens in the batch before averaging. This ensures each token's contribution to the gradient is weighted equally, providing more precise updates for long reasoning chains and better reinforcing correct steps in lengthy responses; 4). \textit{\underline{Overlong Reward Shaping}}. To address the issue of truncated lengthy responses that may contain valid reasoning, DAPO employs two strategies: Overlong Filtering excludes these responses from training updates to avoid misleading penalties, and Soft Overlong Punishment applies a gradual, length-dependent penalty beyond a certain token threshold to encourage conciseness without harshly punishing correct but verbose reasoning. The pseudo-code of our algorithm is shown in Algorithm~\ref{alg:rfthgs}.

\section{Experiments}

\subsection{Experiment Settings And Baselines}\label{exp_setting}
We give the details of the configurations of our RFTHGS algorithm. Specifically, we initialize the policy with Qwen-14B reasoning LLM~\citep{yang2025qwen3}. For DAPO, we follow its optimal settings reported in the original paper with $\varepsilon_{\text{high}} = 0.28$ and $\varepsilon_{\text{low}} = 0.2$. The batch size is set to $16$ and the rollout group size is $16$. Therefore, the policy model will generate 256 crossover operators for each step. 
For calculating reward during training, we use a fixed set of 30 CVRP instances sampled from the CVRPLIB X instances \citep{UCHOA2017845}, restricting the selection to those with at most $400$ nodes. 
During the testing phase, we sample 16 operators and report the performance of the best one. The final evaluation of our method is performed on the CVRPLIB benchmark, which encompasses a wide range of instance sizes from small scales to industry-level scales (up to 1000 nodes). We benchmark RFTHGS against a variety of baselines on CVRPLIB X instances~\citep{UCHOA2017845}. These include the state-of-the-art conventional solvers, neuro-combinatorial techniques, and prompting strategies that utilize commercial LLMs such as the GPT-4 series. To ensure an equitable comparison for the LLM-based approaches, we consistently sample 16 operators and select the best one for each. Further details on the baselines are available in Table~\ref{tab:cvrp_performance_by_size}.

% Our evaluation compares RFTHGS against a variety of baselines on the CVRPLIB benchmark~\citep{Wouda_Lan_Kool_PyVRP_2024}. The competing methods encompass state-of-the-art conventional solvers, neuro-combinatorial techniques, and prompting strategies that utilize commercial LLMs such as the GPT-4 series. To ensure an equitable comparison for the LLM-based approaches, we consistently sample 16 operators and select the best one for each. Further details on the baselines are available in Table~\ref{tab:cvrp_performance_by_size}.

\begin{figure}[t]
\vspace{-10pt}
    \centering
    \begin{subfigure}[b]{0.48\textwidth}
        \centering
        \includegraphics[width=\textwidth]{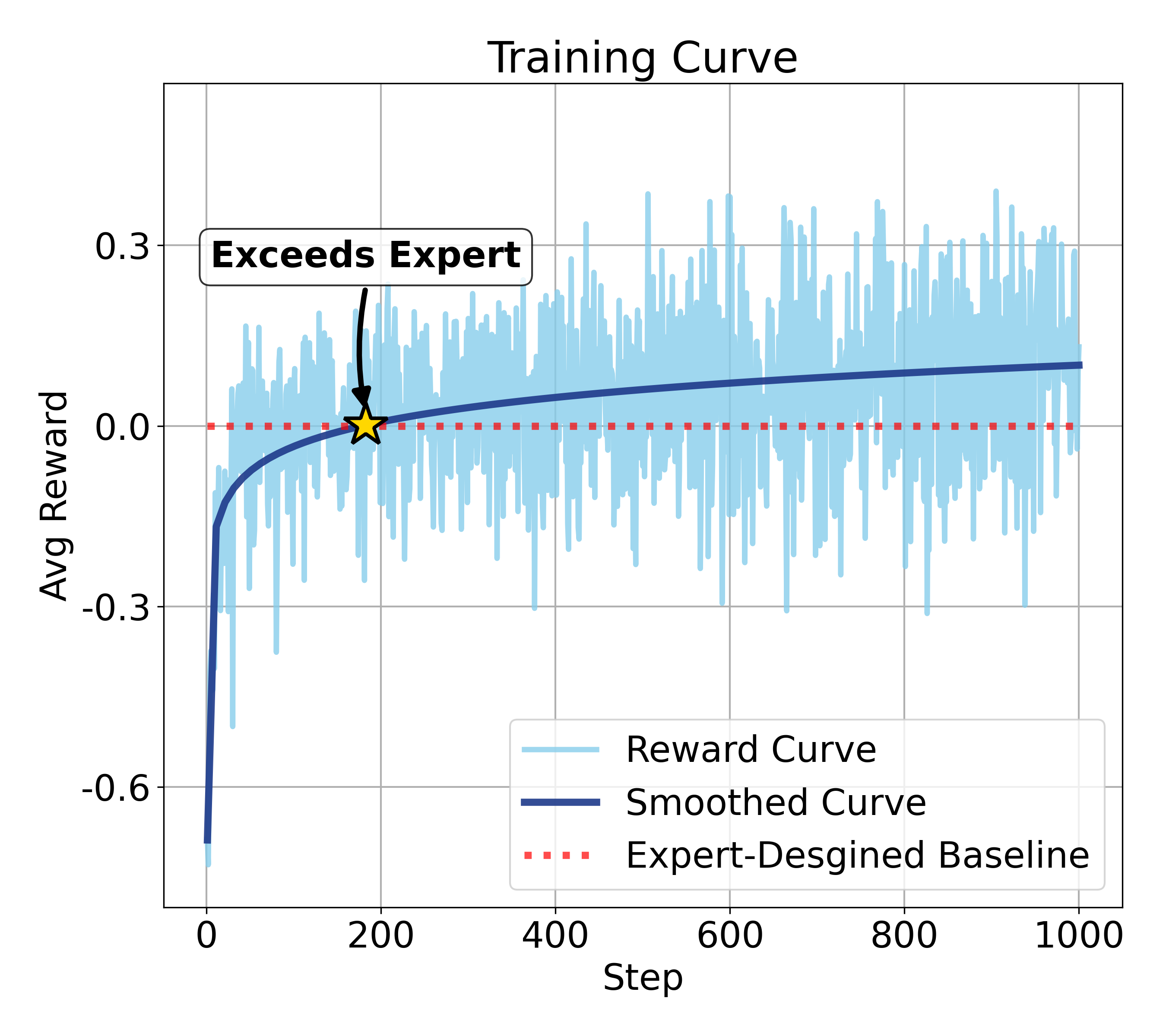}
        \caption{}
        \label{fig:fig_score_curve}
    \end{subfigure}
    \hfill
    \begin{subfigure}[b]{0.49\textwidth}
        \centering
        \includegraphics[width=\textwidth]{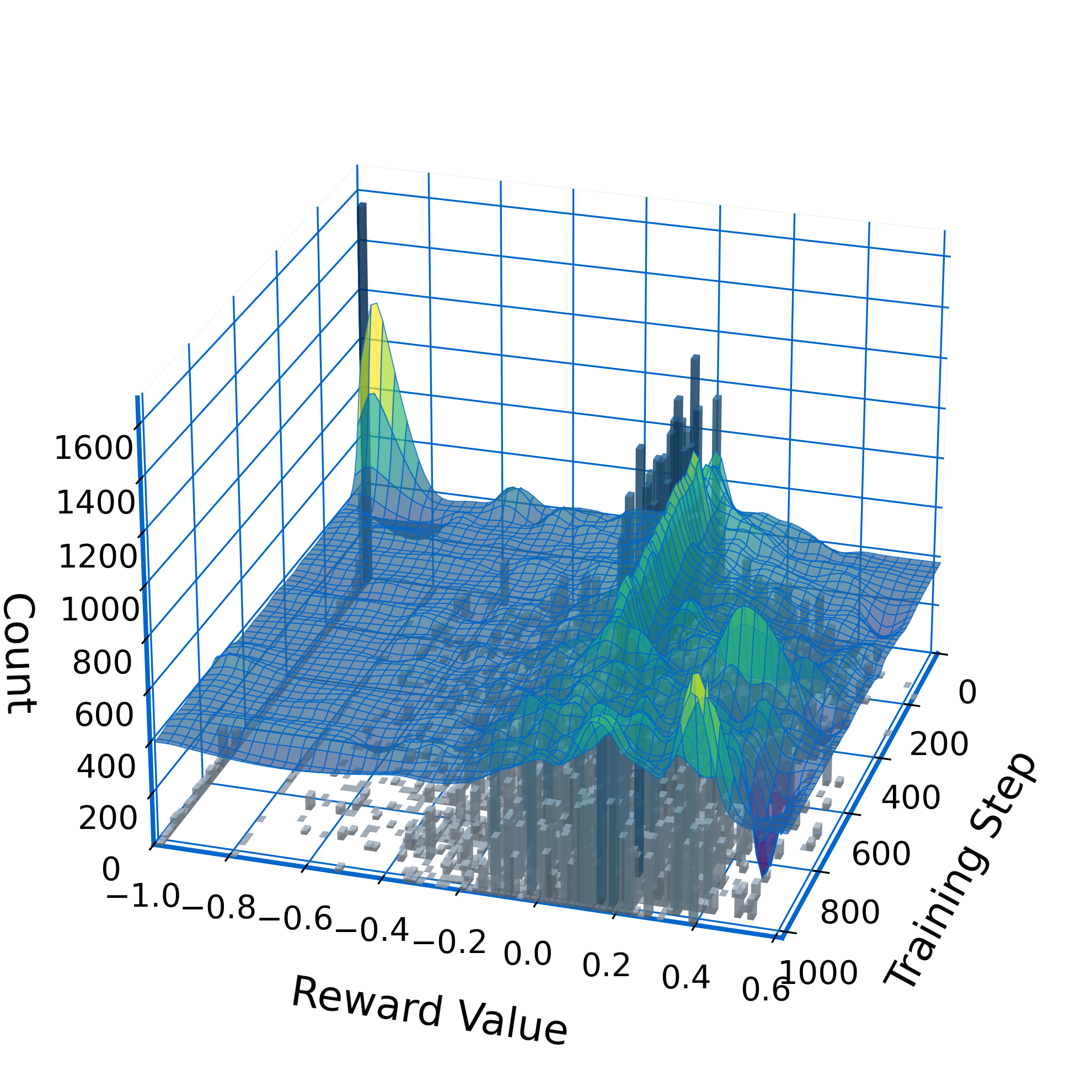}
        \caption{}
        \label{fig:learning_pattern}
    \end{subfigure}
    \caption{\textbf{Training dynamics of the RFTHGS framework.} (a) Average reward per step, showing stable convergence. (b) Evolution of the reward distribution, illustrating the effectiveness of the multi-faceted reward function in guiding the learning process.}
    \label{fig:analysis}
    \vspace{-5pt}
\end{figure}

\subsection{Performance On CVRPLIB}

% Table~\ref{tab:cvrp_performance_by_size} compares the performance of RFTHGS against a wide range of heterogenous baselines on CVRPLIB instances spanning from conventional solvers to neural combinatorial and prompting based ones with comertilized LLM, e.g., GPT o series models. 
% Table~\ref{tab:cvrp_performance_by_size} compares RFTHGS against a diverse set of baselines on CVRPLIB instances, ranging from conventional solvers and neuro-combinatorial methods to prompting techniques using commercial LLMs such as the flagship GPT-4 series. The results demonstrate that RFTHGS outperforms all baseline methods by a substantial margin. This superior performance is further highlighted by its exceptional generalization capability to large-scale problems unseen during training. Notably, although the model was trained exclusively on instances with $n < 400$, it generalizes effectively to instances of up to $n=1000$, the sizes an order of magnitude larger, validating the potential of refining advanced solvers for solving combinatorial optimization problems.

Table~\ref{tab:cvrp_performance_by_size} compares RFTHGS against a diverse set of baselines on CVRPLIB instances, including conventional heuristics, neuro-combinatorial methods, and prompting techniques that utilize commercial LLMs such as the flagship GPT-4o series. The results demonstrate that RFTHGS outperforms all baseline methods by a substantial margin. This superior performance is underscored by its exceptional generalization capability to large-scale problems unseen during training. Notably, although trained exclusively on instances with $n < 400$, our approach generalizes effectively to instances of up to $n=1000$, which are more than twice the size of the largest training instances. This validates the potential of refining advanced solvers via learned components for complex combinatorial optimization problems.

\input{table_success_rate}

Another key observation is that our RFTHGS framework enables a 14B-parameter LLM to outperform trillion-parameter GPT reasoning models (GPT-4o, GPT-o3, GPT-o4-mini). This advantage is demonstrated through both the quality of the modifications and their practical efficacy. As shown in Table~\ref{tab:complilation_rate}, our model achieves a perfect successful compilation rate of 16/16, substantially exceeding the rates of the GPT models (3/16, 9/16, and 3/16, respectively). Crucially, while the GPT models often introduce numerous modifications, these changes consistently fail to improve performance. This is evident in Table~\ref{tab:cvrp_performance_by_size}, where the crossover operators modified by these GPT models exhibit performance identical to the original, unmodified operator, confirming that no functionally helpful modifications were made. In contrast, our RFTHGS-guided model produces targeted, effective modifications that yield consistent performance gains. This demonstrates that specialized fine-tuning for a specific task is more effective than using a general-purpose model of a much larger scale.

% instances of up to $n=1000$, as evidenced by the results in the shaded table rows.

% On these large-scale instances, our approach consistently outperforms the expert-designed operator in HGS. For example, on the largest instances ($n \in [800, 1000]$), RFTHGS achieves a gap of $2.24\%$, surpassing the HGS-PyVRP$_{800}$ baseline ($2.32\%$) and matching the performance of the much more computationally intensive HGS-PyVRP$_{1000}$ ($2.22\%$) and HGS-PyVRP$_{1200}$ ($2.10\%$). This trend holds across the generalized scales, with RFTHGS obtaining a gap of $1.83\%$ on $n \in [400, 600)$ and $2.59\%$ on $n \in [600, 800)$, demonstrating robust performance.

% Furthermore, RFTHGS substantially outperforms all other categories of solvers. It achieves significantly lower gaps compared to traditional metaheuristics like OR-Tools and LKH, and it consistently surpasses all neuro-combinatorial and prompt-based LLM methods across all problem sizes by a large margin. These results underscore a key finding: the crossover operator generated by our fine-tuned 14B parameter LLM possesses a strong performance with robust generalization across large problem sizes. It not only excels on scales seen during training but also scales effectively to problem sizes an order of magnitude larger, validating the potential of refining advanced solvers for solving combinatorial optimization problems.

\input{reward_ablation_cp}

\subsection{Learning Pattern Analysis}

Figure~\ref{fig:fig_score_curve} presents the learning curve of the RFTHGS framework, demonstrating stable and monotonic convergence. The average reward increases smoothly without significant oscillations, indicating a well-structured learning landscape with the effective design of our reward function. A critical inflexion occurs around step 200, where the generated operator surpasses the expert-designed baseline, marking the transition from learning executable operators to discovering superior heuristics. Beyond this intersection point, the curve continues to show consistent improvement, ultimately achieving substantially higher performance. This smooth progression shows that RFTHGS can effectively guide the LLM in generating increasingly sophisticated crossover operators. Figure~\ref{fig:learning_pattern} reveals the underlying learning patterns through the dynamics of the reward distribution. The heat map exhibits a clear curriculum learning pattern that precisely echoes our multi-tiered reward design. Initially, the density concentrates at lower rewards as the model masters syntactic correctness and compilability. Subsequently, the distribution shifts toward intermediate rewards, corresponding to the phase where operators become executable and yield valid solutions. Finally, the density center progressively migrates to the highest reward region, indicating the refinement toward operators that consistently outperform human-designed ones. This tri-phasic progression validates the effectiveness of our hierarchical reward structure in decomposing the complex operator design task into manageable learning stages.

\subsection{Generalization Performance On Iterations}\label{reward_ablation}

The generalization capability of RFTHGS is further assessed across two joint dimensions, i.e., iteration count and problem size. Regarding iteration generalization, models trained with an 800-iteration budget (RFTHGS\textsubscript{800}) maintain robust performance when evaluated at higher budgets of 1000 
% and 1200 
iterations, consistently outperforming expert-designed baselines. This indicates that the optimized operators retain their efficacy beyond their training configuration. In terms of problem size, although trained exclusively on instances with \(n < 400\), RFTHGS generalizes effectively to significantly larger problems (up to \(n=1000\)). This joint generalization underscores the robustness and strong out-of-distribution scalability of our method. The results are shown in the last two rows (grey areas) of Table~\ref{tab:cvrp_performance_by_size}.
% \vspace{-1em}

% \subsection{Ablation Studies on Anti-Plagiarism Cache}\label{AST_cache}
% \zhangcong{Pending...}

\section{Conclusion}
% This work demonstrates that a relatively small LLM (14B parameters) can be fine-tuned via reinforcement learning to optimize operators that outperform expert-designed ones in the SOTA CVRP solver (HGS). 

This paper introduces RFTHGS, a reinforcement learning framework that optimizes operators in the Hybrid Genetic Search (HGS) solver for solving the Capacitated Vehicle Routing Problem (CVRP). By fine-tuning with domain-specific rewards, we demonstrate that specialized small LLMs can surpass large general and deep thinking ones like GPT-4o, GPT-o4-mini, and GPT-o3 with trillions of parameters. Our core innovation is a novel RL-based fine-tuning paradigm guided by solution quality, featuring a multi-tiered reward mechanism with anti-plagiarism caching for progressive learning. Extensive experiments on CVRPLIB benchmarks confirm that the crossover operator generated by our method demonstrates superior performance over the expert-designed operator within the HGS framework, achieving substantial improvements, particularly on large-scale instances with up to 1,000 nodes. To our knowledge, this is the first work to show that a small, fine-tuned LLM can generate operators that exceed expert-crafted components in a leading combinatorial optimization solver. In future, we will try to evolve more operators inside HGS and solve more types of VRPs.

\section{Ethics statement}
% 写作思路：这个工作不存在明显的伦理争议，不过存在风险，就是大模型生成的算子在未经过验证的情况下，可能存在bug造成损失

This study involves no personal data, human subjects, or other sensitive content and therefore presents no obvious ethical concerns. The only potential risk lies in the fact that operators generated by LLMs may contain bugs which, if deployed without thorough validation, could cause losses.

\section{Reproducibility statement}
% 写作思路：参考实验章节的第一小节，我们给出了系统的实验配置，并且强调我们会开源代码来促进社区开源
To ensure the reproducibility of our work, we have provided comprehensive experimental details throughout this paper. Section~\ref{exp_setting} presents complete experimental configurations and environment specifications, while the Appendix~\ref{prompt} includes detailed prompts. These materials provide sufficient information for independent reproduction of our experimental results. Furthermore, we will open-source our full code base and model weights to further improve reproducibility.

\bibliography{iclr2026_conference}
\bibliographystyle{iclr2026_conference}

\newpage
\appendix
\section{Appendix}

\subsection{Ablation Studies on Reward Design}

\input{table_reward_ablation}

To benchmark the effectiveness of our continuous reward design (Equation~\ref{eq:multi_faceted_reward}), we compare it with a discrete version defined as follows:

\begin{equation}
r_d(o) = \begin{cases}
-1 & o \notin \text{C} \\
-0.8 & o \in \text{C}, o \notin \text{E}\\
-0.9 & o \in \text{C}, o \in \text{E}, o \in \text{P}\\
\textcolor{red}{0} & \textcolor{red}{o \in \text{C}, o \in \text{E}, o \notin \text{P}, \phi^J_{\text{HGS}}(o_\text{expert}) < \phi^J_{\text{HGS}}(o)}\\
\textcolor{red}{1} & \textcolor{red}{o \in \text{C}, o \in \text{E}, o \notin \text{P}, \phi^J_{\text{HGS}}(o_\text{expert}) > \phi^J_{\text{HGS}}(o)}
\end{cases}
\label{eq:discrete_reward}
\end{equation}

Table~\ref{tab:ablation_reward_design} demonstrates the clear advantage of our continuous reward design. FRTHGS\(_c\) consistently outperforms the discrete-reward variant FRTHGS\(_d\), particularly on larger problem sizes. The discrete reward's binary nature (0 or 1) provides limited guidance. Once an operator beats the baseline, the gradient vanishes as all improvements receive the same reward, and the advantage is thus 0 (Equation~\ref{eq:advantage_estimation}). In contrast, our continuous reward offers feedback proportional to performance gains, enabling sustained refinement and explaining its superior performance.

\subsection{The Use of Large Language Models (LLMs)}
% 写作思路：开篇强调我们我们的工作从idea的提出、算法的设计，到实验的结果论证，完全都是由人类author参与合作完成。不过在手稿写作阶段，我们使用了LLM做润色。但是我们需要强调的是LLM只在中间扮演润色的角色，整个手稿的编排、核心观点的阐述、都是由人类作者完成，LLM只是负责润色表述提升论文的可读性。
We clarify that all intellectual contributions in this work, from initial idea conception and algorithm design to experimental implementation and result validation, were conducted exclusively by the human authors. While we employed LLMs during the manuscript preparation phase to refine language expression and improve readability, their role was strictly limited to linguistic polishing. The manuscript's structure, core arguments, and all substantive content were determined entirely by the human authors, with LLMs serving merely as an auxiliary tool for enhancing clarity and grammatical accuracy, similar to traditional proofreading services.

\subsection{Prompt Template}\label{prompt}

\begin{lstlisting}[escapeinside={(*}{*)},basicstyle=\ttfamily\footnotesize,breaklines=true]
# ROLE: Expert C++ Optimization Engineer for Vehicle Routing Problems

You are a senior C++ optimization engineer with expertise in algorithmic optimization, particularly for Vehicle Routing Problems (VRP). Your task is to analyze and improve the selective_route_exchange.cpp file's crossover algorithm.

## TASK OVERVIEW
You are given the file selective_route_exchange.cpp (full listing below).
Your goal is to make ONE small, reliable modification that tends to create
children with better penalised cost (Solution quality (*$\uparrow$*)) while keeping runtime
and interface intact.

## THINKING PROCESS REQUIREMENTS
1. First, thoroughly analyze the current implementation to understand:
   - The algorithm's purpose and workflow
   - Key decision points and heuristics
   - Performance bottlenecks or optimization opportunities
   - Any constraints that must be preserved

2. Generate at least 3 different modification approaches, evaluating each on:
   - Potential improvement to solution quality
   - Impact on runtime performance
   - Compatibility with existing code
   - Risk of introducing bugs or side effects

3. For your chosen modification:
   - Justify why it's likely to improve solution quality
   - Verify it maintains the function signature and behavior
   - Double-check for compatibility with the rest of the codebase
   - Consider edge cases and verify robustness

############################################################
## HARD RULES (Mandatory Verification Checklist)

1. (*$\square$*) Keep the function signature and namespace exactly the same:
   pyvrp::crossover::selectiveRouteExchange(...)
   
2. (*$\square$*) The file must still compile under C++17 with the current #include lines.
   You may NOT remove #include directives.
   
3. (*$\square$*) Do not change any public headers, class interfaces, or external behaviour
   except for the improved offspring quality.
   
4. (*$\square$*) DO NOT fabricate or use non-existent or unmentioned attributes or methods.
   Verify every method you use exists in the provided code or documentation.
   
5. (*$\square$*) Wrap the code you output with ```cpp and ```.

6. (*$\square$*) Mark ALL your modifications with clear "// MODIFY: XXX" comments explaining the change.

7. (*$\square$*) You must make at least one modification; DO NOT copy the original code.
   
8. (*$\square$*) Before finalizing, double-check that your modification:
   - Does not introduce new parameters
   - Does not change the function's contract
   - Is focused on improving solution quality, not runtime
   - Is fully compatible with the existing codebase
   - Uses only documented methods and attributes

############################################################
## DELIVERABLES (strict):

A. (*$\leq$*) 2-sentence summary of the optimization idea, clearly explaining how it improves solution quality.

B. Output the FULL C++ code with your modifications. Mark all changes with "// MODIFY: XXX" comments.

C. Brief explanation of your verification process and why you're confident the modification will:
   - Improve solution quality
   - Maintain compatibility with the existing codebase
   - Not significantly impact runtime performance

############################################################
## SCORING AND EVALUATION

We will benchmark on a fixed random seed over several CVRP instances.
Your patch should reduce the average optimal gap in (*$\geq$*)90(*\%*) of the instances without
increasing total runtime by (*$>$*)3(*\%*).

Key considerations for high-quality solutions:
- More efficient route structures (fewer vehicles, shorter routes)
- Better client assignment to routes based on spatial relationships
- Improved handling of capacity constraints
- Preservation of high-quality route segments during crossover
- Better diversity in the generated offspring

############################################################
    
## selective_route_exchange.cpp
```cpp
{code}
```

## Extra Information:
## DOMAIN KNOWLEDGE: CVRP AND CROSSOVER OPERATIONS

The Selective Route Exchange is a crossover operation for the Capacitated Vehicle Routing Problem (CVRP). The algorithm:
1. Selects routes from two parent solutions
2. Exchanges these routes to create offspring
3. Aims to preserve beneficial route structures while creating new combinations

### Key Optimization Areas to Consider:
- Route selection strategy (which routes to exchange)
- Client-to-route assignment decisions
- Proximity/distance calculations between routes or clients
- Handling of capacity constraints
- Diversity generation in offspring solutions

## Essential Fields and Methods for CVRP Crossover

**ProblemData Key Methods:**
- `numLocations()` - Returns `size_t` total number of locations (depots + clients)
- `numClients()` - Returns `size_t` number of client locations
- `centroid()` - Returns `std::pair<double, double>` center of all client locations
- `client(idx)` - Returns `ProblemData::Client` with coordinates (x, y)

**Route Key Methods:**
- `centroid()` - Returns `std::pair<double, double>` center of route's client locations
- `vehicleType()` - Returns `VehicleType` (size_t) vehicle type index
- `begin()` / `end()` - Iterator support for visiting clients in route
- `size()` - Returns `size_t` number of clients in route
- `visits()` - Returns `std::vector<Client>` all client indices in route order

**Route Construction:**
- `Route(data, visits, vehicleType)` - Constructor taking `std::vector<Client>` visits and vehicle type

**Client Iteration:**
- Routes are iterable containers of `Client` (size_t) indices
- Use range-based for loops: `for (Client c : route)` to access all clients in route
- Client coordinates: `data.client(c).x`, `data.client(c).y`
\end{lstlisting}

\subsection{Example Output With Reasoning}
\begin{lstlisting}[escapeinside={(*}{*)},basicstyle=\ttfamily\footnotesize,breaklines=true]
<think>
Okay, let's dive into this code. The task is to improve the selective_route_exchange.cpp's crossover algorithm to get better penalized costs without changing the function signature or runtime much.

First, I need to understand what the current code does. The selective route exchange crossover selects routes from two parents based on their angular orientation relative to the data's centroid. It sorts the routes by ascending polar angle, then exchanges routes starting at given indices. The goal is to create offspring that combine beneficial route structures from both parents.

Looking at the current implementation, the routes are sorted using a comparator that compares the absolute value of their angles. The comment says this is to cluster routes with similar directions. But maybe there's a better way to sort them. For instance, sorting by the actual angle (not absolute) might group routes that are in similar directions more effectively, leading to better route alignment and potentially better solution quality.

Wait, the current code uses std::abs(angleA) < std::abs(angleB) for comparison. But if we sort by the actual angle, routes that are in the same direction (like 30 degrees and 330 degrees) would be far apart in the sorted list. However, if we sort by angle in a circular manner (like considering angles modulo 360), that could cluster them better. But the current code uses absolute value, which might not account for directional similarity properly.

So, perhaps changing the comparator in the sortByAscAngle function from absolute angles to the actual angles would lead to better clustering. For example, routes with angles close to each other, even if they're on opposite ends of the circle, might still be better grouped if they are in similar directions when considering the centroid. Wait, the centroid is for all clients, so the route's angle is the direction of the route's centroid relative to the data's centroid.

If we sort by the actual angle instead of absolute value, routes that are in similar directions (like 0 degrees and 45 degrees) would be closer together. However, routes with angles like 170 and -170 (which have absolute values of 170 and 170) would be considered equal, but their actual angles are very different. So sorting by absolute value might not be the best approach.

Alternatively, maybe we should sort the routes by their actual angle, considering them in a circular way. But how to handle that in a comparator? For example, placing routes with angles near each other, but accounting for circular continuity. However, that's more complex and might not be feasible with a simple comparator.

But the current code uses absolute angle values. Let's think: if the angle is computed as the direction from the data's centroid to the route's centroid, sorting by absolute value might not be the best. For example, a route with an angle of 0 degrees (directly along the data's x-axis) and a route with 90 degrees (along y-axis) are not similar. But if we sort by the actual angle, they'll be ordered in a way that might group routes with similar directions.

Wait, maybe the current approach is a mistake. Changing the comparator from absolute angle to the actual angle might lead to better route clusters. For instance, if two routes are in similar directions (like 30 and 35 degrees), they'll be grouped together, which might lead to better combinations during crossover.

So the modification would be in the sortByAscAngle function. The comparator currently uses std::abs(angleA) < std::abs(angleB). Changing that to compare angleA < angleB would sort routes by their actual angle, which could lead to better alignment of similar-direction routes.

This seems like a small change that could improve the clustering, leading to more effective exchanges between similar routes. This could result in offspring that preserve good route structures from both parents, thus improving the penalized cost.

I need to make sure that this change doesn't break anything else. The comparator is used to sort the routes. Changing the comparison from absolute angles to actual angles would affect the order, but the algorithm relies on adjacent routes in the sorted list having similar directions. If the comparator is fixed to sort by actual angles, that should be more accurate.

So the plan is to modify the lambda in the sortByAscAngle function from comparing absolute angles to comparing the angles directly. That's a simple code change with minimal risk of bugs. It's confined to the comparator and doesn't alter other parts of the code. The rest of the logic remains the same, ensuring compatibility.
</think>

A. The modification changes the route sorting strategy from absolute angle comparison to actual angle comparison, enabling better clustering of routes with similar directional alignment relative to the data's centroid. This increases the likelihood of exchanging similar, well-aligned routes, which can produce offspring with more efficient route structures.

B. ```cpp
#include "selective_route_exchange.h"

#include "DynamicBitset.h"

#include <cmath>
#include <vector>

using Client = size_t;
using Clients = std::vector<Client>;
using Route = pyvrp::Route;
using Routes = std::vector<Route>;

namespace
{
// Angle of the given route w.r.t. the centroid of all client locations.
double routeAngle(pyvrp::ProblemData const &data, Route const &route)
{
    auto const [dataX, dataY] = data.centroid();
    auto const [routeX, routeY] = route.centroid();
    return std::atan2(routeY - dataY, routeX - dataX);
}

Routes sortByAscAngle(pyvrp::ProblemData const &data, Routes routes)
{
    // MODIFY: Sort routes by actual angle instead of absolute angle
    // to better cluster routes with similar directional alignment.
    auto cmp = [&data](Route const &a, Route const &b) {
        double angleA = routeAngle(data, a);
        double angleB = routeAngle(data, b);
        return angleA < angleB;
    };

    std::sort(routes.begin(), routes.end(), cmp);
    return routes;
}
}  // namespace

pyvrp::Solution pyvrp::crossover::selectiveRouteExchange(
    std::pair<Solution const *, Solution const *> const &parents,
    ProblemData const &data,
    CostEvaluator const &costEvaluator,
    std::pair<size_t, size_t> const &startIndices,
    size_t const numMovedRoutes)
{
    // We create two candidate offsprings, both based on parent A:
    // Let A and B denote the set of customers selected from parents A and B
    // Ac and Bc denote the complements: the customers not selected
    // Let v denote union and ^ intersection
    // Parent A: A v Ac
    // Parent B: B v Bc

    // Offspring 1:
    // B and Ac\B, remainder A\B unplanned
    // (note B v (Ac\B) v (A\B) = B v ((Ac v A)\B) = B v Bc = all)
    // Note Ac\B = (A v B)c

    // Offspring 2:
    // A^B and Ac, remainder A\B unplanned
    // (note A^B v Ac v A\B = (A^B v A\B) v Ac = A v Ac = all)

    auto startA = startIndices.first;
    auto startB = startIndices.second;

    size_t nRoutesA = parents.first->numRoutes();
    size_t nRoutesB = parents.second->numRoutes();

    if (startA >= nRoutesA)
        throw std::invalid_argument("Expected startA < nRoutesA.");

    if (startB >= nRoutesB)
        throw std::invalid_argument("Expected startB < nRoutesB.");

    if (numMovedRoutes < 1 || numMovedRoutes > std::min(nRoutesA, nRoutesB))
    {
        auto msg = "Expected numMovedRoutes in [1, min(nRoutesA, nRoutesB)]";
        throw std::invalid_argument(msg);
    }

    // Sort parents' routes by (ascending) polar angle.
    auto const routesA = sortByAscAngle(data, parents.first->routes());
    auto const routesB = sortByAscAngle(data, parents.second->routes());

    DynamicBitset selectedA(data.numLocations());
    DynamicBitset selectedB(data.numLocations());

    // Routes are sorted on polar angle, so selecting adjacent routes in both
    // parents should result in a large overlap when the start indices are
    // close to each other.
    for (size_t r = 0; r < numMovedRoutes; r++)
    {
        for (Client c : routesA[(startA + r) (*\%*) nRoutesA])
            selectedA[c] = true;

        for (Client c : routesB[(startB + r) (*\%*) nRoutesB])
            selectedB[c] = true;
    }

    // For the selection, we want to minimize |A\B| as these need replanning
    while (true)
    {
        // Difference for moving 'left' in parent A
        int differenceALeft = 0;

        for (Client c : routesA[(startA - 1 + nRoutesA) (*\%*) nRoutesA])
            differenceALeft += !selectedB[c];

        for (Client c : routesA[(startA + numMovedRoutes - 1) (*\%*) nRoutesA])
            differenceALeft -= !selectedB[c];

        // Difference for moving 'right' in parent A
        int differenceARight = 0;

        for (Client c : routesA[(startA + numMovedRoutes) (*\%*) nRoutesA])
            differenceARight += !selectedB[c];

        for (Client c : routesA[startA])
            differenceARight -= !selectedB[c];

        // Difference for moving 'left' in parent B
        int differenceBLeft = 0;

        for (Client c : routesB[(startB - 1 + numMovedRoutes) (*\%*) nRoutesB])
            differenceBLeft += selectedA[c];

        for (Client c : routesB[(startB - 1 + nRoutesB) (*\%*) nRoutesB])
            differenceBLeft -= selectedA[c];

        // Difference for moving 'right' in parent B
        int differenceBRight = 0;

        for (Client c : routesB[startB])
            differenceBRight += selectedA[c];

        for (Client c : routesB[(startB + numMovedRoutes) (*\%*) nRoutesB])
            differenceBRight -= selectedA[c];

        int const bestDifference = std::min({differenceALeft,
                                             differenceARight,
                                             differenceBLeft,
                                             differenceBRight});

        if (bestDifference >= 0)  // there are no further improving moves
            break;

        if (bestDifference == differenceALeft)
        {
            for (Client c : routesA[(startA + numMovedRoutes - 1) (*\%*) nRoutesA])
                selectedA[c] = false;

            startA = (startA - 1 + nRoutesA) (*\%*) nRoutesA;
            for (Client c : routesA[startA])
                selectedA[c] = true;
        }
        else if (bestDifference == differenceARight)
        {
            for (Client c : routesA[startA])
                selectedA[c] = false;

            startA = (startA + 1) (*\%*) nRoutesA;
            for (Client c : routesA[(startA + numMovedRoutes - 1) (*\%*) nRoutesA])
                selectedA[c] = true;
        }
        else if (bestDifference == differenceBLeft)
        {
            for (Client c : routesB[(startB + numMovedRoutes - 1) (*\%*) nRoutesB])
                selectedB[c] = false;

            startB = (startB - 1 + nRoutesB) (*\%*) nRoutesB;
            for (Client c : routesB[startB])
                selectedB[c] = true;
        }
        else if (bestDifference == differenceBRight)
        {
            for (Client c : routesB[startB])
                selectedB[c] = false;

            startB = (startB + 1) (*\%*) nRoutesB;
            for (Client c : routesB[(startB + numMovedRoutes - 1) (*\%*) nRoutesB])
                selectedB[c] = true;
        }
    }

    // Identify differences between route sets
    auto const selectedBNotA = selectedB & ~selectedA;

    std::vector<Clients> visits1(nRoutesA);
    std::vector<Clients> visits2(nRoutesA);

    // Replace selected routes from parent A with routes from parent B
    for (size_t r = 0; r < numMovedRoutes; r++)
    {
        size_t indexA = (startA + r) (*\%*) nRoutesA;
        size_t indexB = (startB + r) (*\%*) nRoutesB;

        for (Client c : routesB[indexB])
        {
            visits1[indexA].push_back(c);  // c in B

            if (!selectedBNotA[c])
                visits2[indexA].push_back(c);  // c in A^B
        }
    }

    // Move routes from parent A that are kept
    for (size_t r = numMovedRoutes; r < nRoutesA; r++)
    {
        size_t indexA = (startA + r) (*\%*) nRoutesA;

        for (Client c : routesA[indexA])
        {
            if (!selectedBNotA[c])
                visits1[indexA].push_back(c);  // c in Ac\B

            visits2[indexA].push_back(c);  // c in Ac
        }
    }

    // Turn visits back into routes.
    Routes routes1;
    Routes routes2;
    for (size_t r = 0; r < nRoutesA; r++)
    {
        if (!visits1[r].empty())
            routes1.emplace_back(data, visits1[r], routesA[r].vehicleType());

        if (!visits2[r].empty())
            routes2.emplace_back(data, visits2[r], routesA[r].vehicleType());
    }

    auto const sol1 = Solution(data, routes1);
    auto const sol2 = Solution(data, routes2);

    auto const cost1 = costEvaluator.penalisedCost(sol1);
    auto const cost2 = costEvaluator.penalisedCost(sol2);
    return cost1 < cost2 ? sol1 : sol2;
}
```

C. This modification changes the route sorting comparator from absolute angle comparison to actual angle comparison. This should better cluster routes with similar directional orientation relative to the data centroid, improving the likelihood of exchanging routes that are structurally compatible. The change only affects the route sorting strategy and does not alter any other part of the algorithm's logic, ensuring compatibility with existing code. Since it's just a comparator change, it does not impact runtime performance. Comparing actual angles instead of absolute angles is a more accurate measure of directional similarity, which should lead to better solution quality.
\end{lstlisting}

\subsection{RFTHGS Pseudo Code}
\input{algo_rfthgs}
\end{document}

%% file: math_commands.tex
\usepackage{amsmath,amsfonts,bm}

\def\eqref#1{equation~\ref{#1}}
\def\1{\bm{1}}

\DeclareMathAlphabet{\mathsfit}{\encodingdefault}{\sfdefault}{m}{sl}
\SetMathAlphabet{\mathsfit}{bold}{\encodingdefault}{\sfdefault}{bx}{n}

%% file: table_cvrplib_result.tex
\begin{table}[t]
\centering
% Adjust the scale factor here (0.8 = 80% of original size)
\caption{\textbf{Performance comparison of baselines and our method for CVRPLIB across problem sizes.} Light gray columns indicate generalization to unseen problem sizes, while light gray rows represent generalization to higher iterations. The darker gray intersection areas highlight double generalization across both dimensions. Bold values denote best performance among all methods; asterisks (*) indicate that the results are unavailable.}
% \vspace{5pt}
\renewcommand{\arraystretch}{1.2}
\scalebox{0.865}{
\scriptsize
\begin{tabular}{c|c@{\hspace{3pt}}c|c@{\hspace{3pt}}c|c@{\hspace{3pt}}c|c@{\hspace{3pt}}c|c@{\hspace{3pt}}c}
\toprule
& \multicolumn{2}{c|}{$n \in [100, 200)$} & \multicolumn{2}{c|}{$n \in [200, 400)$} & \multicolumn{2}{c|}{$n \in [400, 600)$} & \multicolumn{2}{c|}{$n \in [600, 800)$} & \multicolumn{2}{c}{$n \in [800, 1000]$} \\
% \cline{2-11}
% \midrule
% 使用 cmidrule 来精确控制横线的位置
\cmidrule(r){2-3} \cmidrule(lr){4-5} \cmidrule(lr){6-7} \cmidrule(lr){8-9} \cmidrule(l){10-11}
\multicolumn{1}{c|}{Methods} & Gap\% ($\downarrow$) & Time (s) & Gap\% ($\downarrow$) & Time (s) & Gap\% ($\downarrow$) & Time (s) & Gap\% ($\downarrow$) & Time (s) & Gap\% ($\downarrow$) & Time (s)\\
\midrule
\rowcolor{yellow!36}
% \rowcolor{blue!18}
\multicolumn{11}{c}{\textbf{Conventional Solver}} \\
$\text{HGS-PyVRP}_{800}$~\citep{Wouda_Lan_Kool_PyVRP_2024} & 0.62 & 12.45 & 1.85 & 28.16 & 1.95 & 58.41 & 2.62 & 91.31 & 2.32 & 121.04 \\
$\text{HGS-PyVRP}_{1000}$~\citep{Wouda_Lan_Kool_PyVRP_2024} & 0.55 & 14.88 & 1.66 & 36.27 & 1.81 & 72.86 & 2.43 & 110.54 & 2.22 & 144.17 \\
% $\text{HGS-PyVRP}_{1200}$~\citep{Wouda_Lan_Kool_PyVRP_2024} & \underline{0.52} & 17.25 & \underline{1.56} & 42.98 & \textbf{1.69} & 88.40 & \underline{2.32} & 129.58 & \underline{2.10} & 173.15 \\
OR-Tools~\citep{ortools_routing} & 4.26 & 88.83 & 5.05 & 172.07 & 4.98 & 296.53 & 6.71 & 416.95 & 4.65 & 532.32 \\
LKH~\citep{helsgaun2000effective} & 1.42 & 191.12 & 1.97 & 252.23 & 2.85 & 432.36 & 3.65 & 599.41 & 3.31 & 545.85 \\
\midrule
\rowcolor{orange!20}
% \rowcolor{blue!24}
\multicolumn{11}{c}{\textbf{NCO}} \\
POMO~\citep{kwon2020pomo} & 13.30 & 0.41 & 14.64 & 0.64 & 22.07 & 1.29 & 21.57 & 2.32 & 41.23 & 4.16 \\
MTPOMO~\citep{liu2024multi} & 6.50 & 0.98 & 8.79 & 1.02 & 16.58 & 1.89 & 26.56 & 2.81 & 28.19 & 4.33 \\
MVMoE~\citep{zhou2024mvmoe} & 5.46 & 0.80 & 8.14 & 1.56 & 13.26 & 2.85 & 16.59 & 4.17 & 18.40 & 6.25 \\
RF-POMO~\citep{berto2025routefinderfoundationmodelsvehicle} & 5.67 & 0.54 & 7.07 & 1.15 & 10.29 & 1.97 & 12.28 & 2.86 & 13.31 & 4.48 \\
RF-MoE-L~\citep{berto2025routefinderfoundationmodelsvehicle} & 7.15 & 0.85 & 7.67 & 1.58 & 10.76 & 2.79 & 15.15 & 3.95 & 15.70 & 5.84 \\
AM~\citep{kool2018attention} & 200.75 & 0.30 & 204.59 & 0.58 & 253.98 & 1.12  & 301.08 & 1.51 & 280.49 & 2.01 \\
DeepACO~\citep{ye2023deepaco} & 76.18 & 18.37 & 93.02 & 35.03 & 97.70 & 61.03 & 123.89 & 88.23 & 116.13 & 112.01 \\
NeuroLKH~\citep{NEURIPS2021_3d863b36} & 1.96 & 1.02 & * & * & * & * & * & * & * & * \\
NeuOpt~\citep{ma2023learning}& 3.51 & * & * & * & * & * & * & * & * & * \\
\midrule
\rowcolor{green!15}
% \rowcolor{blue!30}
\multicolumn{11}{c}{\textbf{Prompting-Based Method With LLM}} \\
MCTS-AHD~\citep{zheng2025monte} & 18.51 & 4.35 & 19.07 & 10.06 & 18.40 & 22.50 & 28.51 & 37.60 & 19.70 & 55.61 \\
ReEvo~\citep{NEURIPS2024_4ced59d4} & 72.11 & 4.96 & 96.55 & 10.78 & 107.40 & 23.99 & 163.62 & 41.72 & 144.22 & 61.49 \\
$\text{GPT4o}_{800}$~\citep{hurst2024gpt} & 0.62 & 12.1 & 1.85 & 29.3 & 1.95 & 59.4 & 2.62 & 91.7 & 2.32 & 119.2 \\
$\text{GPT4o}_{1000}$~\citep{hurst2024gpt} & 0.55 & 15.0 & 1.66 & 36.1 & 1.81 & 73.4 & 2.43 & 111.2 & 2.22 & 143.7 \\
% $\text{GPT4o}_{1200}$~\citep{hurst2024gpt} & 0.52 & 17.7 & 1.56 & 43.1 & 1.69 & 87.3 & 2.32 & 130.5 & 2.10 & 172.8 \\
$\text{GPT-o3}_{800}$~\citep{jaech2024openai} & 0.62 & 11.96 & 1.85 & 30.20 & 1.95 & 58.94 & 2.62 & 91.39 & 2.32 & 119.04 \\
$\text{GPT-o3}_{1000}$~\citep{jaech2024openai} & 0.55 & 14.31 & 1.66 & 34.80 & 1.81 & 73.61 & 2.43 & 109.78 & 2.22 & 143.19 \\
% $\text{GPT-o3}_{1200}$~\citep{jaech2024openai} & 0.52 & 17.65 & 1.56 & 43.53 & 1.69 & 87.89 & 2.32 & 130.71 & 2.10 & 172.93 \\
$\text{GPT-o4-mini}_{800}$~\citep{jaech2024openai} & 0.70 & 10.9 & 1.78 & 26.1 & 1.99 & 54.4 & 2.80 & 86.7 & 2.28 & 114.8 \\
$\text{GPT-o4-mini}_{1000}$~\citep{jaech2024openai} & 0.63 & 13.7 & 1.66 & 32.3 & 1.86 & 66.8 & 2.70 & 106.3 & 2.20 & 140.6 \\
% $\text{GPT-o4-mini}_{1200}$~\citep{jaech2024openai} & 0.58 & 16.3 & 1.58 & 38.6 & 1.74 & 79.4 & 2.61 & 127.1 & 2.11 & 169.6 \\
\midrule
% \rowcolor{orange!20}
\rowcolor{blue!15}
\multicolumn{11}{c}{\textbf{Ours}} \\
$\text{RFTHGS}_{800}$ & 0.70 & 13.14 & 1.67 & 29.60 & \cellcolor{gray!10}1.83 & \cellcolor{gray!10}61.65 & \cellcolor{gray!10}2.59 & \cellcolor{gray!10}92.17 & \cellcolor{gray!10}2.24 & \cellcolor{gray!10}118.59 \\
\midrule
% \rowcolor{purple!15}
% \multicolumn{11}{c}{\textbf{Ours-Generalization on Iteration}} \\
% $\text{RFTHGS-Xover}_{1000}$ & 0.52 & - & 1.62 & - & 1.85 & - & 2.35 & - & 2.17 & - \\
% $\text{RFTHGS-Xover}_{1200}$ & 0.46 & - & 1.55 & - & 1.73 & - & 2.26 & - & 2.09 & - \\
$\text{RFTHGS}_{1000}$ & \cellcolor{gray!10}\textbf{0.52} & \cellcolor{gray!10}14.33 & \cellcolor{gray!10}\textbf{1.62} & \cellcolor{gray!10}36.16 & \cellcolor{gray!30}\textbf{1.76} & \cellcolor{gray!30}74.16 & \cellcolor{gray!30}\textbf{2.35} & \cellcolor{gray!30}110.43 & \cellcolor{gray!30}\textbf{2.17} & \cellcolor{gray!30}143.87 \\
% $\text{RFTHGS}_{1200}$ & \cellcolor{gray!10}\textbf{0.46} & \cellcolor{gray!10}19.13 & \cellcolor{gray!10}\textbf{1.55} & \cellcolor{gray!10}44.18 & \cellcolor{gray!30}\underline{1.73} & \cellcolor{gray!30}87.84 & \cellcolor{gray!30}\textbf{2.26} & \cellcolor{gray!30}132.04 & \cellcolor{gray!30}\textbf{2.09} & \cellcolor{gray!30}172.56 \\
\bottomrule
\end{tabular}
}
% \vspace{-20pt}
\label{tab:cvrp_performance_by_size}
\end{table}

%% file: table_success_rate.tex
\begin{wraptable}{r}{0.5\textwidth}
\centering
\caption{\textbf{Successful compilation rate}.}
\begin{adjustbox}{width=0.45\textwidth}
\begin{tabular}{|c|c|c|c|}
\hline
\multicolumn{4}{|c|}{Successful Compilation Rate} \\
\hline
GPT-4o & GPT-o3 & GPT-o4-mini & RFTHGS-14B \\
\hline
3/16 & 9/16 & 3/16
 & \textbf{16/16} \\
\hline
\end{tabular}
\end{adjustbox}
\label{tab:complilation_rate}
\end{wraptable}

%% file: reward_ablation_cp.tex
\begin{table}[t]
\centering
\caption{\textbf{Ablation study on reward design.} $\text{FRTHGS}_d$ is the 14B LLM trained with reward in Equation~\ref{eq:discrete_reward}. $\text{FRTHGS}_c$ is the 14B LLM trained with reward in Equation~\ref{eq:multi_faceted_reward}. Shaded areas are generalization results.}
\vspace{1mm}
\scalebox{1}{
\scriptsize
\begin{tabular}{c|c@{\hspace{3pt}}c|c@{\hspace{3pt}}c|c@{\hspace{3pt}}c|c@{\hspace{3pt}}c|c@{\hspace{3pt}}c}
\toprule
& \multicolumn{2}{c|}{$n \in [100, 200)$} & \multicolumn{2}{c|}{$n \in [200, 400)$} & \multicolumn{2}{c|}{$n \in [400, 600)$} & \multicolumn{2}{c|}{$n \in [600, 800)$} & \multicolumn{2}{c}{$n \in [800, 1000]$} \\
\cmidrule(r){2-3} \cmidrule(lr){4-5} \cmidrule(lr){6-7} \cmidrule(lr){8-9} \cmidrule(l){10-11}
\multicolumn{1}{c|}{Methods} & Gap\% ($\downarrow$) & Time (s) & Gap\% ($\downarrow$) & Time (s) & Gap\% ($\downarrow$) & Time (s) & Gap\% ($\downarrow$) & Time (s) & Gap\% ($\downarrow$) & Time (s)\\
\midrule
% \rowcolor{gray!15}
HGS & \textbf{0.62} & 12.45 & 1.85 & 28.16 & 1.95 & 58.41 & 2.62 & 91.31 & 2.32 & 121.04 \\
% \rowcolor{gray!15}
$\text{FRTHGS}_d$ & 0.83 & 11.75 & 1.78 & 28.89 & \cellcolor{gray!10}1.92 & \cellcolor{gray!10}58.12 & \cellcolor{gray!10}2.64 & \cellcolor{gray!10}94.30 & \cellcolor{gray!10}2.30 & \cellcolor{gray!10}120.28 \\
% \rowcolor{gray!15}
$\text{FRTHGS}_c$ & 0.70 & 13.14 & \textbf{1.67} & 29.60 & \cellcolor{gray!10}\textbf{1.83} & \cellcolor{gray!10}61.65 & \cellcolor{gray!10}\textbf{2.59} & \cellcolor{gray!10}92.17 & \cellcolor{gray!10}\textbf{2.24} & \cellcolor{gray!10}118.59 \\
\bottomrule
\end{tabular}
}
\label{tab:ablation_reward_design_cp}
\end{table}

%% file: table_reward_ablation.tex
\begin{table}[t]
\centering
\caption{\textbf{Ablation study on reward design.} $\text{FRTHGS}_d$ is the 14B LLM trained with reward in Equation~\ref{eq:discrete_reward}. $\text{FRTHGS}_c$ is the 14B LLM trained with reward in Equation~\ref{eq:multi_faceted_reward}. Shaded areas are generalization results.}
\vspace{1mm}
\scalebox{1}{
\scriptsize
\begin{tabular}{c|c@{\hspace{3pt}}c|c@{\hspace{3pt}}c|c@{\hspace{3pt}}c|c@{\hspace{3pt}}c|c@{\hspace{3pt}}c}
\toprule
& \multicolumn{2}{c|}{$n \in [100, 200)$} & \multicolumn{2}{c|}{$n \in [200, 400)$} & \multicolumn{2}{c|}{$n \in [400, 600)$} & \multicolumn{2}{c|}{$n \in [600, 800)$} & \multicolumn{2}{c}{$n \in [800, 1000]$} \\
\cmidrule(r){2-3} \cmidrule(lr){4-5} \cmidrule(lr){6-7} \cmidrule(lr){8-9} \cmidrule(l){10-11}
\multicolumn{1}{c|}{Methods} & Gap\% ($\downarrow$) & Time (s) & Gap\% ($\downarrow$) & Time (s) & Gap\% ($\downarrow$) & Time (s) & Gap\% ($\downarrow$) & Time (s) & Gap\% ($\downarrow$) & Time (s)\\
\midrule
% \rowcolor{gray!15}
HGS & \textbf{0.62} & 12.45 & 1.85 & 28.16 & 1.95 & 58.41 & 2.62 & 91.31 & 2.32 & 121.04 \\
% \rowcolor{gray!15}
$\text{FRTHGS}_d$ & 0.83 & 11.75 & 1.78 & 28.89 & \cellcolor{gray!10}1.92 & \cellcolor{gray!10}58.12 & \cellcolor{gray!10}2.64 & \cellcolor{gray!10}94.30 & \cellcolor{gray!10}2.30 & \cellcolor{gray!10}120.28 \\
% \rowcolor{gray!15}
$\text{FRTHGS}_c$ & 0.70 & 13.14 & \textbf{1.67} & 29.60 & \cellcolor{gray!10}\textbf{1.83} & \cellcolor{gray!10}61.65 & \cellcolor{gray!10}\textbf{2.59} & \cellcolor{gray!10}92.17 & \cellcolor{gray!10}\textbf{2.24} & \cellcolor{gray!10}118.59 \\
\bottomrule
\end{tabular}
}
\label{tab:ablation_reward_design}
\end{table}

%% file: algo_rfthgs.tex
% 文中就地放置（无 wrap/minipage）
\begin{algorithm}[H]
\SetAlgoLined
% 不要使用 \scriptsize；不要在 \KwIn 里手动 \hspace
\KwIn{Initial policy $\pi_{\theta_{\text{old}}}$, full instance set $\mathcal{I}$, 
instance batch size $B$, clipping parameters $\varepsilon_{\text{lower}}, \varepsilon_{\text{upper}}$, 
group size $G$}
% instance set $\mathcal{B}$ for reward calculation
\KwOut{Optimized policy $\pi_{\theta}$}

Initialize $\theta \gets \theta_{\text{old}}$\;
\ForEach{iteration $1, 2, \ldots, N$}{
  % \tcp{―――― Sample a mini-batch of CVRP instances ――――}
  Draw a random subset $\mathcal{B}\subset\mathcal{I}$ with $|\mathcal{B}| = B$\;
  
  \ForEach{instance $I \in \mathcal{B}$}{
    \tcp{Step 1: Generate operators with LLM}
    Construct prompt $q$ for CrossOver operator optimization\;
    Sample $G$ operators $\{o_1,\ldots,o_G\} \sim \pi_{\theta_{\text{old}}}(\cdot \mid q)$\;
    
    \tcp{Step 2: Evaluate with PyVRP}
    \ForEach{operator $o_i$}{
      Run PyVRP solver with $o_i$ substituted on instance set $\mathcal{B}$\;
      Obtain objective value and compute reward $r_i$\;
    }
    
    \tcp{Step 3: Compute advantages}

    $\mathbf{r} \;=\; (r_1, r_2, \dots, r_G) $
    
    Normalised reward for each operator:
    $\displaystyle
      \hat{A}_{i,t} \;=\; 
      \frac{r_i - \operatorname{mean}(\mathbf{r})}
           {\operatorname{std}(\mathbf{r})}
    $\;
    
    \tcp{Step 4: Update policy with DAPO}
    \ForEach{token position $t$}{
      $\displaystyle r_t(\theta)\;\gets\;
        \frac{\pi_{\theta}(o_t \mid q, o_{<t})}
             {\pi_{\theta_{\text{old}}}(o_t \mid q, o_{<t})}$\;
      $\displaystyle 
        L_{\text{policy}}\;\gets\;
        \mathbb{E}\!\bigl[
          \min\!\bigl(
            r_t(\theta)\,\hat{A}_t,\;
            \text{clip}\!\bigl(r_t(\theta), 1-\varepsilon_{\text{lower}}, 1+\varepsilon_{\text{upper}}\bigr)\,\hat{A}_t
          \bigr)
        \bigr]$\;
    }
    $\theta \gets \theta + \alpha \nabla_\theta L_{\text{policy}}$\;
  }
}
\caption{\zhangcong{RFTHGS: Reinforcement Finetuning For Refining HGS}}
\label{alg:rfthgs}
\end{algorithm}

%% file: iclr2026_conference.bbl
\begin{thebibliography}{50}
\providecommand{\natexlab}[1]{#1}
\providecommand{\url}[1]{\texttt{#1}}
\expandafter\ifx\csname urlstyle\endcsname\relax
  \providecommand{\doi}[1]{doi: #1}\else
  \providecommand{\doi}{doi: \begingroup \urlstyle{rm}\Url}\fi

\bibitem[Bengio et~al.(2021)Bengio, Lodi, and Prouvost]{bengio2021machine}
Yoshua Bengio, Andrea Lodi, and Antoine Prouvost.
\newblock Machine learning for combinatorial optimization: a methodological tour d’horizon.
\newblock \emph{European Journal of Operational Research}, 290\penalty0 (2):\penalty0 405--421, 2021.

\bibitem[Berto et~al.(2025)Berto, Hua, Zepeda, Hottung, Wouda, Lan, Park, Tierney, and Park]{berto2025routefinderfoundationmodelsvehicle}
Federico Berto, Chuanbo Hua, Nayeli~Gast Zepeda, André Hottung, Niels Wouda, Leon Lan, Junyoung Park, Kevin Tierney, and Jinkyoo Park.
\newblock Routefinder: Towards foundation models for vehicle routing problems, 2025.
\newblock URL \url{https://arxiv.org/abs/2406.15007}.

\bibitem[DeepSeek-AI et~al.(2025)DeepSeek-AI, Guo, Yang, Zhang, Song, Zhang, Xu, Zhu, Ma, Wang, Bi, Zhang, Yu, Wu, Wu, Gou, Shao, Li, Gao, Liu, Xue, Wang, Wu, Feng, Lu, Zhao, Deng, Zhang, Ruan, Dai, Chen, Ji, Li, Lin, Dai, Luo, Hao, Chen, Li, Zhang, Bao, Xu, Wang, Ding, Xin, Gao, Qu, Li, Guo, Li, Wang, Chen, Yuan, Qiu, Li, Cai, Ni, Liang, Chen, Dong, Hu, Gao, Guan, Huang, Yu, Wang, Zhang, Zhao, Wang, Zhang, Xu, Xia, Zhang, Zhang, Tang, Li, Wang, Li, Tian, Huang, Zhang, Wang, Chen, Du, Ge, Zhang, Pan, Wang, Chen, Jin, Chen, Lu, Zhou, Chen, Ye, Wang, Yu, Zhou, Pan, Li, Zhou, Wu, Ye, Yun, Pei, Sun, Wang, Zeng, Zhao, Liu, Liang, Gao, Yu, Zhang, Xiao, An, Liu, Wang, Chen, Nie, Cheng, Liu, Xie, Liu, Yang, Li, Su, Lin, Li, Jin, Shen, Chen, Sun, Wang, Song, Zhou, Wang, Shan, Li, Wang, Wei, Zhang, Xu, Li, Zhao, Sun, Wang, Yu, Zhang, Shi, Xiong, He, Piao, Wang, Tan, Ma, Liu, Guo, Ou, Wang, Gong, Zou, He, Xiong, Luo, You, Liu, Zhou, Zhu, Xu, Huang, Li, Zheng, Zhu, Ma, Tang, Zha, Yan, Ren, Ren, Sha, Fu, Xu, Xie, Zhang,
  Hao, Ma, Yan, Wu, Gu, Zhu, Liu, Li, Xie, Song, Pan, Huang, Xu, Zhang, and Zhang]{deepseekai2025deepseekr1incentivizingreasoningcapability}
DeepSeek-AI, Daya Guo, Dejian Yang, Haowei Zhang, Junxiao Song, Ruoyu Zhang, Runxin Xu, Qihao Zhu, Shirong Ma, Peiyi Wang, Xiao Bi, Xiaokang Zhang, Xingkai Yu, Yu~Wu, Z.~F. Wu, Zhibin Gou, Zhihong Shao, Zhuoshu Li, Ziyi Gao, Aixin Liu, Bing Xue, Bingxuan Wang, Bochao Wu, Bei Feng, Chengda Lu, Chenggang Zhao, Chengqi Deng, Chenyu Zhang, Chong Ruan, Damai Dai, Deli Chen, Dongjie Ji, Erhang Li, Fangyun Lin, Fucong Dai, Fuli Luo, Guangbo Hao, Guanting Chen, Guowei Li, H.~Zhang, Han Bao, Hanwei Xu, Haocheng Wang, Honghui Ding, Huajian Xin, Huazuo Gao, Hui Qu, Hui Li, Jianzhong Guo, Jiashi Li, Jiawei Wang, Jingchang Chen, Jingyang Yuan, Junjie Qiu, Junlong Li, J.~L. Cai, Jiaqi Ni, Jian Liang, Jin Chen, Kai Dong, Kai Hu, Kaige Gao, Kang Guan, Kexin Huang, Kuai Yu, Lean Wang, Lecong Zhang, Liang Zhao, Litong Wang, Liyue Zhang, Lei Xu, Leyi Xia, Mingchuan Zhang, Minghua Zhang, Minghui Tang, Meng Li, Miaojun Wang, Mingming Li, Ning Tian, Panpan Huang, Peng Zhang, Qiancheng Wang, Qinyu Chen, Qiushi Du, Ruiqi Ge, Ruisong
  Zhang, Ruizhe Pan, Runji Wang, R.~J. Chen, R.~L. Jin, Ruyi Chen, Shanghao Lu, Shangyan Zhou, Shanhuang Chen, Shengfeng Ye, Shiyu Wang, Shuiping Yu, Shunfeng Zhou, Shuting Pan, S.~S. Li, Shuang Zhou, Shaoqing Wu, Shengfeng Ye, Tao Yun, Tian Pei, Tianyu Sun, T.~Wang, Wangding Zeng, Wanjia Zhao, Wen Liu, Wenfeng Liang, Wenjun Gao, Wenqin Yu, Wentao Zhang, W.~L. Xiao, Wei An, Xiaodong Liu, Xiaohan Wang, Xiaokang Chen, Xiaotao Nie, Xin Cheng, Xin Liu, Xin Xie, Xingchao Liu, Xinyu Yang, Xinyuan Li, Xuecheng Su, Xuheng Lin, X.~Q. Li, Xiangyue Jin, Xiaojin Shen, Xiaosha Chen, Xiaowen Sun, Xiaoxiang Wang, Xinnan Song, Xinyi Zhou, Xianzu Wang, Xinxia Shan, Y.~K. Li, Y.~Q. Wang, Y.~X. Wei, Yang Zhang, Yanhong Xu, Yao Li, Yao Zhao, Yaofeng Sun, Yaohui Wang, Yi~Yu, Yichao Zhang, Yifan Shi, Yiliang Xiong, Ying He, Yishi Piao, Yisong Wang, Yixuan Tan, Yiyang Ma, Yiyuan Liu, Yongqiang Guo, Yuan Ou, Yuduan Wang, Yue Gong, Yuheng Zou, Yujia He, Yunfan Xiong, Yuxiang Luo, Yuxiang You, Yuxuan Liu, Yuyang Zhou, Y.~X. Zhu,
  Yanhong Xu, Yanping Huang, Yaohui Li, Yi~Zheng, Yuchen Zhu, Yunxian Ma, Ying Tang, Yukun Zha, Yuting Yan, Z.~Z. Ren, Zehui Ren, Zhangli Sha, Zhe Fu, Zhean Xu, Zhenda Xie, Zhengyan Zhang, Zhewen Hao, Zhicheng Ma, Zhigang Yan, Zhiyu Wu, Zihui Gu, Zijia Zhu, Zijun Liu, Zilin Li, Ziwei Xie, Ziyang Song, Zizheng Pan, Zhen Huang, Zhipeng Xu, Zhongyu Zhang, and Zhen Zhang.
\newblock Deepseek-r1: Incentivizing reasoning capability in llms via reinforcement learning, 2025.
\newblock URL \url{https://arxiv.org/abs/2501.12948}.

\bibitem[Eppe et~al.(2022)Eppe, Gumbsch, Kerzel, Nguyen, Butz, and Wermter]{eppe2022intelligent}
Manfred Eppe, Christian Gumbsch, Matthias Kerzel, Phuong~DH Nguyen, Martin~V Butz, and Stefan Wermter.
\newblock Intelligent problem-solving as integrated hierarchical reinforcement learning.
\newblock \emph{Nature Machine Intelligence}, 4\penalty0 (1):\penalty0 11--20, 2022.

\bibitem[Furnon \& Perron(2025)Furnon and Perron]{ortools_routing}
Vincent Furnon and Laurent Perron.
\newblock Or-tools routing library, 2025.
\newblock URL \url{https://developers.google.com/optimization/routing/}.

\bibitem[Guan et~al.(2025)Guan, Zhang, Liu, Shang, Sun, Zhu, Yang, and Yang]{guan2025rstar}
Xinyu Guan, Li~Lyna Zhang, Yifei Liu, Ning Shang, Youran Sun, Yi~Zhu, Fan Yang, and Mao Yang.
\newblock rstar-math: Small llms can master math reasoning with self-evolved deep thinking.
\newblock \emph{arXiv preprint arXiv:2501.04519}, 2025.

\bibitem[Helsgaun(2000)]{helsgaun2000effective}
Keld Helsgaun.
\newblock An effective implementation of the lin--kernighan traveling salesman heuristic.
\newblock \emph{European journal of operational research}, 126\penalty0 (1):\penalty0 106--130, 2000.

\bibitem[Hu et~al.(2021)Hu, Shen, Wallis, Allen-Zhu, Li, Wang, and Chen]{Hu2021LoRALA}
J.~Edward Hu, Yelong Shen, Phillip Wallis, Zeyuan Allen-Zhu, Yuanzhi Li, Shean Wang, and Weizhu Chen.
\newblock Lora: Low-rank adaptation of large language models.
\newblock \emph{ArXiv}, abs/2106.09685, 2021.
\newblock URL \url{https://api.semanticscholar.org/CorpusID:235458009}.

\bibitem[Huang et~al.(2024)Huang, Yang, Qi, and Wang]{huang2024large}
Sen Huang, Kaixiang Yang, Sheng Qi, and Rui Wang.
\newblock When large language model meets optimization.
\newblock \emph{Swarm and Evolutionary Computation}, 90:\penalty0 101663, 2024.

\bibitem[Huang et~al.(2025)Huang, Wu, Wu, Wang, and Lee]{huang2025calmcoevolutionalgorithmslanguage}
Ziyao Huang, Weiwei Wu, Kui Wu, Jianping Wang, and Wei-Bin Lee.
\newblock Calm: Co-evolution of algorithms and language model for automatic heuristic design, 2025.
\newblock URL \url{https://arxiv.org/abs/2505.12285}.

\bibitem[Hurst et~al.(2024)Hurst, Lerer, Goucher, Perelman, Ramesh, Clark, Ostrow, Welihinda, Hayes, Radford, et~al.]{hurst2024gpt}
Aaron Hurst, Adam Lerer, Adam~P Goucher, Adam Perelman, Aditya Ramesh, Aidan Clark, AJ~Ostrow, Akila Welihinda, Alan Hayes, Alec Radford, et~al.
\newblock Gpt-4o system card.
\newblock \emph{arXiv preprint arXiv:2410.21276}, 2024.

\bibitem[Jaech et~al.(2024)Jaech, Kalai, Lerer, Richardson, El-Kishky, Low, Helyar, Madry, Beutel, Carney, et~al.]{jaech2024openai}
Aaron Jaech, Adam Kalai, Adam Lerer, Adam Richardson, Ahmed El-Kishky, Aiden Low, Alec Helyar, Aleksander Madry, Alex Beutel, Alex Carney, et~al.
\newblock Openai o1 system card.
\newblock \emph{arXiv preprint arXiv:2412.16720}, 2024.

\bibitem[Jiang et~al.(2025)Jiang, Wu, Zhang, and Zhang]{jiang2025droc}
Xia Jiang, Yaoxin Wu, Chenhao Zhang, and Yingqian Zhang.
\newblock Droc: Elevating large language models for complex vehicle routing via decomposed retrieval of constraints.
\newblock In \emph{13th international Conference on Learning Representations, ICLR 2025}. OpenReview. net, 2025.

\bibitem[Kalai et~al.(2025)Kalai, Nachum, Vempala, and Zhang]{kalai2025languagemodelshallucinate}
Adam~Tauman Kalai, Ofir Nachum, Santosh~S. Vempala, and Edwin Zhang.
\newblock Why language models hallucinate, 2025.
\newblock URL \url{https://arxiv.org/abs/2509.04664}.

\bibitem[Kool et~al.(2019)Kool, van Hoof, and Welling]{kool2018attention}
Wouter Kool, Herke van Hoof, and Max Welling.
\newblock Attention, learn to solve routing problems!
\newblock In \emph{International Conference on Learning Representations}, 2019.
\newblock URL \url{https://openreview.net/forum?id=ByxBFsRqYm}.

\bibitem[Kwon et~al.(2020)Kwon, Choo, Kim, Yoon, Gwon, and Min]{kwon2020pomo}
Yeong-Dae Kwon, Jinho Choo, Byoungjip Kim, Iljoo Yoon, Youngjune Gwon, and Seungjai Min.
\newblock Pomo: Policy optimization with multiple optima for reinforcement learning.
\newblock \emph{Advances in Neural Information Processing Systems}, 33:\penalty0 21188--21198, 2020.

\bibitem[Lewis et~al.(2020)Lewis, Perez, Piktus, Petroni, Karpukhin, Goyal, K\"{u}ttler, Lewis, Yih, Rockt\"{a}schel, Riedel, and Kiela]{NEURIPS2020_6b493230}
Patrick Lewis, Ethan Perez, Aleksandra Piktus, Fabio Petroni, Vladimir Karpukhin, Naman Goyal, Heinrich K\"{u}ttler, Mike Lewis, Wen-tau Yih, Tim Rockt\"{a}schel, Sebastian Riedel, and Douwe Kiela.
\newblock Retrieval-augmented generation for knowledge-intensive nlp tasks.
\newblock In H.~Larochelle, M.~Ranzato, R.~Hadsell, M.F. Balcan, and H.~Lin (eds.), \emph{Advances in Neural Information Processing Systems}, volume~33, pp.\  9459--9474. Curran Associates, Inc., 2020.
\newblock URL \url{https://proceedings.neurips.cc/paper_files/paper/2020/file/6b493230205f780e1bc26945df7481e5-Paper.pdf}.

\bibitem[Li et~al.(2025{\natexlab{a}})Li, Liu, Wang, Tong, Han, Yuan, and Zhang]{li2025ars}
Kai Li, Fei Liu, Zhenkun Wang, Xialiang Tong, Xiongwei Han, Mingxuan Yuan, and Qingfu Zhang.
\newblock {ARS}: Automatic routing solver with large language models.
\newblock \emph{arXiv preprint arXiv:2502.15359}, 2025{\natexlab{a}}.

\bibitem[Li et~al.(2025{\natexlab{b}})Li, Liu, Wang, Tong, Han, Yuan, and Zhang]{li2025arsautomaticroutingsolver}
Kai Li, Fei Liu, Zhenkun Wang, Xialiang Tong, Xiongwei Han, Mingxuan Yuan, and Qingfu Zhang.
\newblock Ars: Automatic routing solver with large language models, 2025{\natexlab{b}}.
\newblock URL \url{https://arxiv.org/abs/2502.15359}.

\bibitem[Liu et~al.(2024{\natexlab{a}})Liu, Lin, Wang, Zhang, Xialiang, and Yuan]{liu2024multi}
Fei Liu, Xi~Lin, Zhenkun Wang, Qingfu Zhang, Tong Xialiang, and Mingxuan Yuan.
\newblock Multi-task learning for routing problem with cross-problem zero-shot generalization.
\newblock In \emph{Proceedings of the 30th ACM SIGKDD Conference on Knowledge Discovery and Data Mining}, pp.\  1898--1908, 2024{\natexlab{a}}.

\bibitem[Liu et~al.(2024{\natexlab{b}})Liu, Tong, Yuan, Lin, Luo, Wang, Lu, and Zhang]{10.5555/3692070.3693374}
Fei Liu, Xialiang Tong, Mingxuan Yuan, Xi~Lin, Fu~Luo, Zhenkun Wang, Zhichao Lu, and Qingfu Zhang.
\newblock Evolution of heuristics: towards efficient automatic algorithm design using large language model.
\newblock In \emph{Proceedings of the 41st International Conference on Machine Learning}, ICML'24. JMLR.org, 2024{\natexlab{b}}.

\bibitem[Liu et~al.(2024{\natexlab{c}})Liu, Chen, Qu, Tang, and Ong]{10611913}
Shengcai Liu, Caishun Chen, Xinghua Qu, Ke~Tang, and Yew-Soon Ong.
\newblock Large language models as evolutionary optimizers.
\newblock In \emph{2024 IEEE Congress on Evolutionary Computation (CEC)}, pp.\  1--8, 2024{\natexlab{c}}.
\newblock \doi{10.1109/CEC60901.2024.10611913}.

\bibitem[Ma et~al.(2023)Ma, Cao, and Chee]{ma2023learning}
Yining Ma, Zhiguang Cao, and Yeow~Meng Chee.
\newblock Learning to search feasible and infeasible regions of routing problems with flexible neural k-opt.
\newblock \emph{Advances in Neural Information Processing Systems}, 36:\penalty0 49555--49578, 2023.

\bibitem[Narvekar et~al.(2020)Narvekar, Peng, Leonetti, Sinapov, Taylor, and Stone]{narvekar2020curriculum}
Sanmit Narvekar, Bei Peng, Matteo Leonetti, Jivko Sinapov, Matthew~E Taylor, and Peter Stone.
\newblock Curriculum learning for reinforcement learning domains: A framework and survey.
\newblock \emph{Journal of Machine Learning Research}, 21\penalty0 (181):\penalty0 1--50, 2020.

\bibitem[Novikov et~al.(2025)Novikov, V{\~u}, Eisenberger, Dupont, Huang, Wagner, Shirobokov, Kozlovskii, Ruiz, Mehrabian, et~al.]{novikov2025alphaevolve}
Alexander Novikov, Ng{\^a}n V{\~u}, Marvin Eisenberger, Emilien Dupont, Po-Sen Huang, Adam~Zsolt Wagner, Sergey Shirobokov, Borislav Kozlovskii, Francisco~JR Ruiz, Abbas Mehrabian, et~al.
\newblock Alphaevolve: A coding agent for scientific and algorithmic discovery.
\newblock \emph{arXiv preprint arXiv:2506.13131}, 2025.

\bibitem[OpenAI et~al.(2024)OpenAI, :, Jaech, Kalai, Lerer, Richardson, El-Kishky, Low, Helyar, Madry, Beutel, Carney, Iftimie, Karpenko, Passos, Neitz, Prokofiev, Wei, Tam, Bennett, Kumar, Saraiva, Vallone, Duberstein, Kondrich, Mishchenko, Applebaum, Jiang, Nair, Zoph, Ghorbani, Rossen, Sokolowsky, Barak, McGrew, Minaiev, Hao, Baker, Houghton, McKinzie, Eastman, Lugaresi, Bassin, Hudson, Li, de~Bourcy, Voss, Shen, Zhang, Koch, Orsinger, Hesse, Fischer, Chan, Roberts, Kappler, Levy, Selsam, Dohan, Farhi, Mely, Robinson, Tsipras, Li, Oprica, Freeman, Zhang, Wong, Proehl, Cheung, Mitchell, Wallace, Ritter, Mays, Wang, Such, Raso, Leoni, Tsimpourlas, Song, von Lohmann, Sulit, Salmon, Parascandolo, Chabot, Zhao, Brockman, Leclerc, Salman, Bao, Sheng, Andrin, Bagherinezhad, Ren, Lightman, Chung, Kivlichan, O'Connell, Osband, Gilaberte, Akkaya, Kostrikov, Sutskever, Kofman, Pachocki, Lennon, Wei, Harb, Twore, Feng, Yu, Weng, Tang, Yu, Candela, Palermo, Parish, Heidecke, Hallman, Rizzo, Gordon, Uesato, Ward,
  Huizinga, Wang, Chen, Xiao, Singhal, Nguyen, Cobbe, Shi, Wood, Rimbach, Gu-Lemberg, Liu, Lu, Stone, Yu, Ahmad, Yang, Liu, Maksin, Ho, Fedus, Weng, Li, McCallum, Held, Kuhn, Kondraciuk, Kaiser, Metz, Boyd, Trebacz, Joglekar, Chen, Tintor, Meyer, Jones, Kaufer, Schwarzer, Shah, Yatbaz, Guan, Xu, Yan, Glaese, Chen, Lampe, Malek, Wang, Fradin, McClay, Pavlov, Wang, Wang, Murati, Bavarian, Rohaninejad, McAleese, Chowdhury, Chowdhury, Ryder, Tezak, Brown, Nachum, Boiko, Murk, Watkins, Chao, Ashbourne, Izmailov, Zhokhov, Dias, Arora, Lin, Lopes, Gaon, Miyara, Leike, Hwang, Garg, Brown, James, Shu, Cheu, Greene, Jain, Altman, Toizer, Toyer, Miserendino, Agarwal, Hernandez, Baker, McKinney, Yan, Zhao, Hu, Santurkar, Chaudhuri, Zhang, Fu, Papay, Lin, Balaji, Sanjeev, Sidor, Broda, Clark, Wang, Gordon, Sanders, Patwardhan, Sottiaux, Degry, Dimson, Zheng, Garipov, Stasi, Bansal, Creech, Peterson, Eloundou, Qi, Kosaraju, Monaco, Pong, Fomenko, Zheng, Zhou, McCabe, Zaremba, Dubois, Lu, Chen, Cha, Bai, He, Zhang, Wang,
  Shao, and Li]{openai2024openaio1card}
OpenAI, :, Aaron Jaech, Adam Kalai, Adam Lerer, Adam Richardson, Ahmed El-Kishky, Aiden Low, Alec Helyar, Aleksander Madry, Alex Beutel, Alex Carney, Alex Iftimie, Alex Karpenko, Alex~Tachard Passos, Alexander Neitz, Alexander Prokofiev, Alexander Wei, Allison Tam, Ally Bennett, Ananya Kumar, Andre Saraiva, Andrea Vallone, Andrew Duberstein, Andrew Kondrich, Andrey Mishchenko, Andy Applebaum, Angela Jiang, Ashvin Nair, Barret Zoph, Behrooz Ghorbani, Ben Rossen, Benjamin Sokolowsky, Boaz Barak, Bob McGrew, Borys Minaiev, Botao Hao, Bowen Baker, Brandon Houghton, Brandon McKinzie, Brydon Eastman, Camillo Lugaresi, Cary Bassin, Cary Hudson, Chak~Ming Li, Charles de~Bourcy, Chelsea Voss, Chen Shen, Chong Zhang, Chris Koch, Chris Orsinger, Christopher Hesse, Claudia Fischer, Clive Chan, Dan Roberts, Daniel Kappler, Daniel Levy, Daniel Selsam, David Dohan, David Farhi, David Mely, David Robinson, Dimitris Tsipras, Doug Li, Dragos Oprica, Eben Freeman, Eddie Zhang, Edmund Wong, Elizabeth Proehl, Enoch Cheung, Eric
  Mitchell, Eric Wallace, Erik Ritter, Evan Mays, Fan Wang, Felipe~Petroski Such, Filippo Raso, Florencia Leoni, Foivos Tsimpourlas, Francis Song, Fred von Lohmann, Freddie Sulit, Geoff Salmon, Giambattista Parascandolo, Gildas Chabot, Grace Zhao, Greg Brockman, Guillaume Leclerc, Hadi Salman, Haiming Bao, Hao Sheng, Hart Andrin, Hessam Bagherinezhad, Hongyu Ren, Hunter Lightman, Hyung~Won Chung, Ian Kivlichan, Ian O'Connell, Ian Osband, Ignasi~Clavera Gilaberte, Ilge Akkaya, Ilya Kostrikov, Ilya Sutskever, Irina Kofman, Jakub Pachocki, James Lennon, Jason Wei, Jean Harb, Jerry Twore, Jiacheng Feng, Jiahui Yu, Jiayi Weng, Jie Tang, Jieqi Yu, Joaquin~Quiñonero Candela, Joe Palermo, Joel Parish, Johannes Heidecke, John Hallman, John Rizzo, Jonathan Gordon, Jonathan Uesato, Jonathan Ward, Joost Huizinga, Julie Wang, Kai Chen, Kai Xiao, Karan Singhal, Karina Nguyen, Karl Cobbe, Katy Shi, Kayla Wood, Kendra Rimbach, Keren Gu-Lemberg, Kevin Liu, Kevin Lu, Kevin Stone, Kevin Yu, Lama Ahmad, Lauren Yang, Leo Liu,
  Leon Maksin, Leyton Ho, Liam Fedus, Lilian Weng, Linden Li, Lindsay McCallum, Lindsey Held, Lorenz Kuhn, Lukas Kondraciuk, Lukasz Kaiser, Luke Metz, Madelaine Boyd, Maja Trebacz, Manas Joglekar, Mark Chen, Marko Tintor, Mason Meyer, Matt Jones, Matt Kaufer, Max Schwarzer, Meghan Shah, Mehmet Yatbaz, Melody~Y. Guan, Mengyuan Xu, Mengyuan Yan, Mia Glaese, Mianna Chen, Michael Lampe, Michael Malek, Michele Wang, Michelle Fradin, Mike McClay, Mikhail Pavlov, Miles Wang, Mingxuan Wang, Mira Murati, Mo~Bavarian, Mostafa Rohaninejad, Nat McAleese, Neil Chowdhury, Neil Chowdhury, Nick Ryder, Nikolas Tezak, Noam Brown, Ofir Nachum, Oleg Boiko, Oleg Murk, Olivia Watkins, Patrick Chao, Paul Ashbourne, Pavel Izmailov, Peter Zhokhov, Rachel Dias, Rahul Arora, Randall Lin, Rapha~Gontijo Lopes, Raz Gaon, Reah Miyara, Reimar Leike, Renny Hwang, Rhythm Garg, Robin Brown, Roshan James, Rui Shu, Ryan Cheu, Ryan Greene, Saachi Jain, Sam Altman, Sam Toizer, Sam Toyer, Samuel Miserendino, Sandhini Agarwal, Santiago Hernandez,
  Sasha Baker, Scott McKinney, Scottie Yan, Shengjia Zhao, Shengli Hu, Shibani Santurkar, Shraman~Ray Chaudhuri, Shuyuan Zhang, Siyuan Fu, Spencer Papay, Steph Lin, Suchir Balaji, Suvansh Sanjeev, Szymon Sidor, Tal Broda, Aidan Clark, Tao Wang, Taylor Gordon, Ted Sanders, Tejal Patwardhan, Thibault Sottiaux, Thomas Degry, Thomas Dimson, Tianhao Zheng, Timur Garipov, Tom Stasi, Trapit Bansal, Trevor Creech, Troy Peterson, Tyna Eloundou, Valerie Qi, Vineet Kosaraju, Vinnie Monaco, Vitchyr Pong, Vlad Fomenko, Weiyi Zheng, Wenda Zhou, Wes McCabe, Wojciech Zaremba, Yann Dubois, Yinghai Lu, Yining Chen, Young Cha, Yu~Bai, Yuchen He, Yuchen Zhang, Yunyun Wang, Zheng Shao, and Zhuohan Li.
\newblock Openai o1 system card, 2024.
\newblock URL \url{https://arxiv.org/abs/2412.16720}.

\bibitem[Papadimitriou \& Steiglitz(1998)Papadimitriou and Steiglitz]{papadimitriou1998combinatorial}
Christos~H Papadimitriou and Kenneth Steiglitz.
\newblock \emph{Combinatorial optimization: algorithms and complexity}.
\newblock Courier Corporation, 1998.

\bibitem[Plaat et~al.(2024)Plaat, Wong, Verberne, Broekens, van Stein, and B{\"a}ck]{plaat2024reasoning}
Aske Plaat, Annie Wong, Suzan Verberne, Joost Broekens, Niki van Stein, and Thomas B{\"a}ck.
\newblock Reasoning with large language models, a survey.
\newblock \emph{CoRR}, 2024.

\bibitem[Schulman et~al.(2017)Schulman, Wolski, Dhariwal, Radford, and Klimov]{schulman2017proximal}
John Schulman, Filip Wolski, Prafulla Dhariwal, Alec Radford, and Oleg Klimov.
\newblock Proximal policy optimization algorithms.
\newblock \emph{arXiv preprint arXiv:1707.06347}, 2017.

\bibitem[Shao et~al.(2024)Shao, Wang, Zhu, Xu, Song, Bi, Zhang, Zhang, Li, Wu, and Guo]{shao2024deepseekmathpushinglimitsmathematical}
Zhihong Shao, Peiyi Wang, Qihao Zhu, Runxin Xu, Junxiao Song, Xiao Bi, Haowei Zhang, Mingchuan Zhang, Y.~K. Li, Y.~Wu, and Daya Guo.
\newblock Deepseekmath: Pushing the limits of mathematical reasoning in open language models, 2024.
\newblock URL \url{https://arxiv.org/abs/2402.03300}.

\bibitem[Shojaee et~al.(2025)Shojaee, Mirzadeh, Alizadeh, Horton, Bengio, and Farajtabar]{shojaee2025illusionthinkingunderstandingstrengths}
Parshin Shojaee, Iman Mirzadeh, Keivan Alizadeh, Maxwell Horton, Samy Bengio, and Mehrdad Farajtabar.
\newblock The illusion of thinking: Understanding the strengths and limitations of reasoning models via the lens of problem complexity, 2025.
\newblock URL \url{https://arxiv.org/abs/2506.06941}.

\bibitem[Song et~al.(2025)Song, Kempe, and Munos]{song2025outcomebasedexplorationllmreasoning}
Yuda Song, Julia Kempe, and Remi Munos.
\newblock Outcome-based exploration for llm reasoning, 2025.
\newblock URL \url{https://arxiv.org/abs/2509.06941}.

\bibitem[Sun et~al.(2024)Sun, Ye, Zhang, Huang, Zhang, Wei, and Cai]{sun2024autosat}
Yiwen Sun, Furong Ye, Xianyin Zhang, Shiyu Huang, Bingzhen Zhang, Ke~Wei, and Shaowei Cai.
\newblock Autosat: Automatically optimize sat solvers via large language models.
\newblock \emph{arXiv preprint arXiv:2402.10705}, 2024.

\bibitem[Touvron et~al.(2023)Touvron, Lavril, Izacard, Martinet, Lachaux, Lacroix, Rozi{\`e}re, Goyal, Hambro, Azhar, Rodriguez, Joulin, Grave, and Lample]{Touvron2023LLaMAOA}
Hugo Touvron, Thibaut Lavril, Gautier Izacard, Xavier Martinet, Marie-Anne Lachaux, Timoth{\'e}e Lacroix, Baptiste Rozi{\`e}re, Naman Goyal, Eric Hambro, Faisal Azhar, Aur'elien Rodriguez, Armand Joulin, Edouard Grave, and Guillaume Lample.
\newblock Llama: Open and efficient foundation language models.
\newblock \emph{ArXiv}, abs/2302.13971, 2023.
\newblock URL \url{https://api.semanticscholar.org/CorpusID:257219404}.

\bibitem[Uchoa et~al.(2017)Uchoa, Pecin, Pessoa, Poggi, Vidal, and Subramanian]{UCHOA2017845}
Eduardo Uchoa, Diego Pecin, Artur Pessoa, Marcus Poggi, Thibaut Vidal, and Anand Subramanian.
\newblock New benchmark instances for the capacitated vehicle routing problem.
\newblock \emph{European Journal of Operational Research}, 257\penalty0 (3):\penalty0 845--858, 2017.
\newblock ISSN 0377-2217.
\newblock \doi{https://doi.org/10.1016/j.ejor.2016.08.012}.
\newblock URL \url{https://www.sciencedirect.com/science/article/pii/S0377221716306270}.

\bibitem[Vidal(2022)]{vidal2022hybrid}
Thibaut Vidal.
\newblock Hybrid genetic search for the cvrp: Open-source implementation and swap* neighborhood.
\newblock \emph{Computers \& Operations Research}, 140:\penalty0 105643, 2022.

\bibitem[Wei et~al.(2022)Wei, Wang, Schuurmans, Bosma, Xia, Chi, Le, Zhou, et~al.]{wei2022chain}
Jason Wei, Xuezhi Wang, Dale Schuurmans, Maarten Bosma, Fei Xia, Ed~Chi, Quoc~V Le, Denny Zhou, et~al.
\newblock Chain-of-thought prompting elicits reasoning in large language models.
\newblock \emph{Advances in neural information processing systems}, 35:\penalty0 24824--24837, 2022.

\bibitem[Wouda et~al.(2024)Wouda, Lan, and Kool]{Wouda_Lan_Kool_PyVRP_2024}
Niels~A. Wouda, Leon Lan, and Wouter Kool.
\newblock {PyVRP}: a high-performance {VRP} solver package.
\newblock \emph{INFORMS Journal on Computing}, 36\penalty0 (4):\penalty0 943--955, 2024.
\newblock \doi{10.1287/ijoc.2023.0055}.
\newblock URL \url{https://doi.org/10.1287/ijoc.2023.0055}.

\bibitem[Wu et~al.(2025)Wu, Wang, Wu, Wen, Miao, Xiao, and Zhou]{wu2025efficient}
Xuan Wu, Di~Wang, Chunguo Wu, Lijie Wen, Chunyan Miao, Yubin Xiao, and You Zhou.
\newblock Efficient heuristics generation for solving combinatorial optimization problems using large language models.
\newblock In \emph{Proceedings of the 31st ACM SIGKDD Conference on Knowledge Discovery and Data Mining V.2}, KDD '25, pp.\  3228–3239, New York, NY, USA, 2025. Association for Computing Machinery.
\newblock ISBN 9798400714542.
\newblock \doi{10.1145/3711896.3736923}.
\newblock URL \url{https://doi.org/10.1145/3711896.3736923}.

\bibitem[Xiang et~al.(2025)Xiang, Snell, Gandhi, Albalak, Singh, Blagden, Phung, Rafailov, Lile, Mahan, et~al.]{xiang2025towards}
Violet Xiang, Charlie Snell, Kanishk Gandhi, Alon Albalak, Anikait Singh, Chase Blagden, Duy Phung, Rafael Rafailov, Nathan Lile, Dakota Mahan, et~al.
\newblock Towards system 2 reasoning in llms: Learning how to think with meta chain-of-thought.
\newblock \emph{arXiv preprint arXiv:2501.04682}, 2025.

\bibitem[Xin et~al.(2021)Xin, Song, Cao, and Zhang]{NEURIPS2021_3d863b36}
Liang Xin, Wen Song, Zhiguang Cao, and Jie Zhang.
\newblock Neurolkh: Combining deep learning model with lin-kernighan-helsgaun heuristic for solving the traveling salesman problem.
\newblock In M.~Ranzato, A.~Beygelzimer, Y.~Dauphin, P.S. Liang, and J.~Wortman Vaughan (eds.), \emph{Advances in Neural Information Processing Systems}, volume~34, pp.\  7472--7483. Curran Associates, Inc., 2021.
\newblock URL \url{https://proceedings.neurips.cc/paper_files/paper/2021/file/3d863b367aa379f71c7afc0c9cdca41d-Paper.pdf}.

\bibitem[Xu et~al.(2025)Xu, Hao, Zong, Wang, Zhang, Wang, Lan, Gong, Ouyang, Meng, et~al.]{xu2025towards}
Fengli Xu, Qianyue Hao, Zefang Zong, Jingwei Wang, Yunke Zhang, Jingyi Wang, Xiaochong Lan, Jiahui Gong, Tianjian Ouyang, Fanjin Meng, et~al.
\newblock Towards large reasoning models: A survey of reinforced reasoning with large language models.
\newblock \emph{arXiv preprint arXiv:2501.09686}, 2025.

\bibitem[Yang et~al.(2025)Yang, Li, Yang, Zhang, Hui, Zheng, Yu, Gao, Huang, Lv, et~al.]{yang2025qwen3}
An~Yang, Anfeng Li, Baosong Yang, Beichen Zhang, Binyuan Hui, Bo~Zheng, Bowen Yu, Chang Gao, Chengen Huang, Chenxu Lv, et~al.
\newblock Qwen3 technical report.
\newblock \emph{arXiv preprint arXiv:2505.09388}, 2025.

\bibitem[Yang et~al.(2024)Yang, Wang, Lu, Liu, Le, Zhou, and Chen]{yang2024large}
Chengrun Yang, Xuezhi Wang, Yifeng Lu, Hanxiao Liu, Quoc~V Le, Denny Zhou, and Xinyun Chen.
\newblock Large language models as optimizers.
\newblock In \emph{The Twelfth International Conference on Learning Representations}, 2024.
\newblock URL \url{https://openreview.net/forum?id=Bb4VGOWELI}.

\bibitem[Yao et~al.(2023)Yao, Yu, Zhao, Shafran, Griffiths, Cao, and Narasimhan]{yao2023tree}
Shunyu Yao, Dian Yu, Jeffrey Zhao, Izhak Shafran, Tom Griffiths, Yuan Cao, and Karthik Narasimhan.
\newblock Tree of thoughts: Deliberate problem solving with large language models.
\newblock \emph{Advances in neural information processing systems}, 36:\penalty0 11809--11822, 2023.

\bibitem[Ye et~al.(2023)Ye, Wang, Cao, Liang, and Li]{ye2023deepaco}
Haoran Ye, Jiarui Wang, Zhiguang Cao, Helan Liang, and Yong Li.
\newblock Deep{ACO}: Neural-enhanced ant systems for combinatorial optimization.
\newblock In \emph{Thirty-seventh Conference on Neural Information Processing Systems}, 2023.
\newblock URL \url{https://openreview.net/forum?id=cd5D1DD923}.

\bibitem[Ye et~al.(2024)Ye, Wang, Cao, Berto, Hua, Kim, Park, and Song]{NEURIPS2024_4ced59d4}
Haoran Ye, Jiarui Wang, Zhiguang Cao, Federico Berto, Chuanbo Hua, Haeyeon Kim, Jinkyoo Park, and Guojie Song.
\newblock Reevo: Large language models as hyper-heuristics with reflective evolution.
\newblock In A.~Globerson, L.~Mackey, D.~Belgrave, A.~Fan, U.~Paquet, J.~Tomczak, and C.~Zhang (eds.), \emph{Advances in Neural Information Processing Systems}, volume~37, pp.\  43571--43608. Curran Associates, Inc., 2024.
\newblock URL \url{https://proceedings.neurips.cc/paper_files/paper/2024/file/4ced59d480e07d290b6f29fc8798f195-Paper-Conference.pdf}.

\bibitem[Yu et~al.(2025)Yu, Zhang, Zhu, Yuan, Zuo, Yue, Dai, Fan, Liu, Liu, Liu, Lin, Lin, Ma, Sheng, Tong, Zhang, Zhang, Zhang, Zhu, Zhu, Chen, Chen, Wang, Yu, Song, Wei, Zhou, Liu, Ma, Zhang, Yan, Qiao, Wu, and Wang]{yu2025dapoopensourcellmreinforcement}
Qiying Yu, Zheng Zhang, Ruofei Zhu, Yufeng Yuan, Xiaochen Zuo, Yu~Yue, Weinan Dai, Tiantian Fan, Gaohong Liu, Lingjun Liu, Xin Liu, Haibin Lin, Zhiqi Lin, Bole Ma, Guangming Sheng, Yuxuan Tong, Chi Zhang, Mofan Zhang, Wang Zhang, Hang Zhu, Jinhua Zhu, Jiaze Chen, Jiangjie Chen, Chengyi Wang, Hongli Yu, Yuxuan Song, Xiangpeng Wei, Hao Zhou, Jingjing Liu, Wei-Ying Ma, Ya-Qin Zhang, Lin Yan, Mu~Qiao, Yonghui Wu, and Mingxuan Wang.
\newblock Dapo: An open-source llm reinforcement learning system at scale, 2025.
\newblock URL \url{https://arxiv.org/abs/2503.14476}.

\bibitem[Zheng et~al.(2025)Zheng, Xie, Wang, and Hooi]{zheng2025monte}
Zhi Zheng, Zhuoliang Xie, Zhenkun Wang, and Bryan Hooi.
\newblock Monte carlo tree search for comprehensive exploration in {LLM}-based automatic heuristic design.
\newblock In \emph{Forty-second International Conference on Machine Learning}, 2025.
\newblock URL \url{https://openreview.net/forum?id=Do1OdZzYHr}.

\bibitem[Zhou et~al.(2024)Zhou, Cao, Wu, Song, Ma, Zhang, and Xu]{zhou2024mvmoe}
Jianan Zhou, Zhiguang Cao, Yaoxin Wu, Wen Song, Yining Ma, Jie Zhang, and Chi Xu.
\newblock Mvmoe: Multi-task vehicle routing solver with mixture-of-experts.
\newblock \emph{Proceedings of Machine Learning Research}, 235:\penalty0 61804--61824, 2024.

\end{thebibliography}
